\documentclass[10pt,twocolumn,letterpaper]{article}

\usepackage[pagenumbers]{wacv}

\usepackage{array}
\usepackage{colortbl}
\usepackage{placeins}
\usepackage{float}
\usepackage{stfloats}

\definecolor{effbg}{HTML}{EAF2FB}
\definecolor{effhead}{HTML}{CFE2F4}
\definecolor{rewbg}{HTML}{EAF5EF}
\definecolor{rewhead}{HTML}{CDE8D8}
\definecolor{fidbg}{HTML}{F1ECF9}
\definecolor{fidhead}{HTML}{DCD0EF}
\definecolor{safebg}{HTML}{FCEBE8}
\definecolor{densebg}{HTML}{F3F3F4}

\definecolor{wacvblue}{rgb}{0.21,0.49,0.74}
\usepackage[pagebackref,breaklinks,colorlinks,allcolors=wacvblue]{hyperref}

\def\wacvPaperID{74} 
\def\confName{WACV}
\def\confYear{2027}

\title{SAFE-DiT: Semantics-Aware Fast-path Execution for High-Resolution Diffusion Transformers}

\author{
Xuanhua Yin\thanks{Equal contribution.} \quad
Yuxuan Jia\footnotemark[1] \quad
Chuanzhi Xu \quad
Weidong Cai\thanks{Corresponding author.}\\
School of Computer Science, The University of Sydney, Australia\\
{\tt\small \{xuanhua.yin, yjia0539, chuanzhi.xu, tom.cai\}@sydney.edu.au}
}

\begin{document}
\maketitle
\begin{abstract}
High-resolution Diffusion Transformer (DiT) inference contains substantial spatial redundancy, but many spatially adaptive implementations encode regional computation as attention masks, which can inadvertently move scaled dot-product attention (SDPA) away from FlashAttention fast paths. We identify this avoidable systems bottleneck as \emph{Mask-Induced Dispatch Tax} (MIDT) and show that it grows with latent sequence length. We introduce SAFE-DiT, a training-free \emph{Semantics-Aware Fast-path Execution} framework that separates exact mask elision from approximation-based spatial scheduling. SAFE-DiT removes only provenance-certified image self-attention masks that induce a row-wise constant shift in attention logits, preserves semantics-bearing masks such as text-padding masks, and realizes spatial adaptation through prompt-conditioned token partitioning, selective state updates with global context, and periodic context refresh. We call this acceleration-only configuration SAFE-Core, and report sensitivity-weighted classifier-free guidance separately as SAFE-DiT+SW. On the evaluated PyTorch SDPA stack, redundant masks make long-sequence attention \(4.1\)--\(5.8\times\) slower than the mask-free path. On Lumina-Next, SAFE-DiT achieves \(2.69\times\) end-to-end acceleration at \(1024^2\) and \(5.09\times\) at \(2560^2\), reduces peak memory at \(2560^2\) from \(94.1\) to \(27.9\) GB, and enables \(3072^2\) generation when dense inference runs out of memory. Paired metrics, component ablations, and a blinded human study support visual non-inferiority of SAFE-Core to the dense fast-path baseline, while SAFE-DiT+SW provides a separate prompt-alignment operating point without reintroducing spatial self-attention masks. Code is available at \url{https://github.com/xuanhuayin/SAFE-DiT}.
\end{abstract}

\section{Introduction}
\label{sec:intro}

High-resolution text-to-image generation needs selective computation: only some regions require frequent updates, yet efficient spatial adaptation in Transformer-based diffusion and flow models remains difficult.

Denoising diffusion models
established iterative generation \cite{ho2020ddpm}, latent diffusion made
synthesis practical in compressed latent spaces \cite{rombach2022ldm}, and
Diffusion Transformers, or DiTs, replaced convolutional denoisers with scalable
token architectures derived from Transformers and Vision Transformers
\cite{vaswani2017attention,dosovitskiy2021vit,peebles2023dit}. Large systems
such as GLIDE~\cite{nichol2022glide}, DALL-E 2~\cite{ramesh2022dalle2},
Imagen~\cite{saharia2022imagen}, eDiff-I~\cite{balaji2022ediffi}, and
SDXL~\cite{podell2024sdxl} established prompt-conditioned synthesis at
scale; recent DiT-style and rectified-flow models, including
PixArt-\(\alpha\)~\cite{chen2023pixartalpha},
PixArt-\(\Sigma\)~\cite{chen2024pixartsigma},
Stable Diffusion 3~\cite{esser2024sd3}, and
Lumina-T2X~\cite{gao2024lumina}, extend prompt following, typography,
and native high-resolution generation. This scaling creates a stringent
inference problem: the image-token sequence grows quadratically with linear
resolution, making repeated attention and block evaluation major sources of
latency and memory pressure.

Spatial adaptation is natural because generation difficulty is not uniform over
the canvas. Prompt-critical objects, text, and fine foreground structures may
need frequent updates, whereas context regions can change slowly across
adjacent denoising steps. This motivates regional prompting, inpainting,
grounded control~\cite{zhang2023controlnet,li2023gligen},
cross-attention intervention~\cite{hertz2023prompttoprompt,tang2023daam,chefer2023attendexcite},
spatial guidance~\cite{bar2023multidiffusion}, and conditional or semantic-aware
guidance strengths~\cite{dhariwal2021guided,ho2022cfg,shen2024scfg}. In parallel,
inference accelerators shorten trajectories~\cite{song2021ddim,lu2022dpmsolver},
distill generation~\cite{song2023consistency}, or reuse features~\cite{ma2024deepcache,liu2024teacache}.
These approaches expose temporal and spatial redundancy, but leave an
orthogonal systems question: how should a spatial decision be
represented so that its algorithmic savings survive attention-kernel dispatch?

\begin{figure*}[t]
  \centering
  \includegraphics[width=0.9\linewidth]{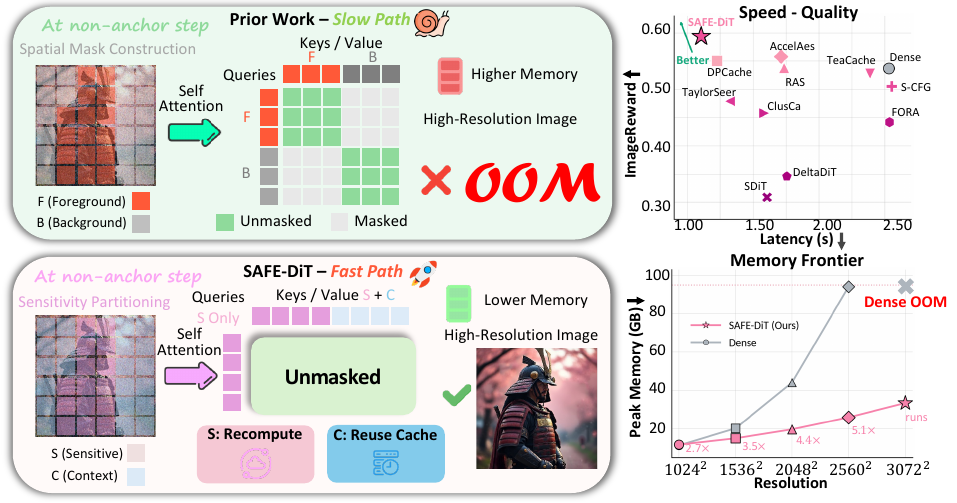}
  \caption{\textbf{SAFE-DiT separates exact mask elision from mask-free spatial
  scheduling.} Certified redundant image self-attention masks are removed to
  avoid the Mask-Induced Dispatch Tax, semantics-bearing text masks are
  retained, and non-anchor computation is expressed through sensitive-query
  updates and context reuse.}
  \label{fig:teaser}
  \vspace{-8pt}
\end{figure*}

The representation is consequential because attention masks can conflate
semantics and routing. Some masks encode padding, causality, blocked
connectivity, or non-uniform attention bias; others are introduced only as
routing metadata, even when they leave the mathematical attention operator
unchanged. Modern inference stacks rely on
FlashAttention-style kernels that avoid materializing the full attention
matrix \cite{dao2022flashattention,dao2023flashattention2}, while PyTorch
scaled dot-product attention, or SDPA, selects among backends according to the
supplied inputs and constraints \cite{pytorch2026sdpa}. Systems such as
FlashMask \cite{wang2025flashmask} and FlexAttention \cite{dong2025flexattention} improve execution when customized masks are
necessary. However, passing a
functionally redundant mask can still select a slower or more memory-intensive
path. We call this avoidable interface cost \emph{\textbf{Mask-Induced Dispatch Tax}},
or \textbf{MIDT}. Under our evaluated stack, paired attention probes show a
\(2.79\)--\(5.91\times\) gap between masked SDPA and the mask-free Flash path
as sequence length grows. MIDT is therefore not an argument against masked
attention; it identifies the narrower case in which a mask is supplied despite
having no semantic effect.

Safe elimination requires an operator-level criterion rather than a mask-type
heuristic. For additive attention, a mask is removable exactly when every query
row adds only a constant shift to all key logits, since row-wise softmax cancels
that shift. All-valid Boolean and all-zero additive masks are special cases;
text padding, causal or block masks, and finite non-uniform biases alter the
normalized attention distribution and must be preserved. This distinction
separates exact mask elision from approximate selective state reuse, whose
speed--fidelity trade-off must be measured independently.

In this paper, we propose \textbf{\underline{SAFE}-DiT}, a
\emph{\textbf{\underline{S}emantics-\underline{A}ware \underline{F}ast-path \underline{E}xecution}}
framework for spatially adaptive high-resolution DiT inference. As illustrated
in Fig.~\ref{fig:teaser}, \textbf{SAFE-DiT} rewrites only masks whose provenance
certifies operator redundancy, then represents spatial adaptation outside the
self-attention mask interface.
\textbf{Prompt-Conditioned Sensitivity Partitioning}, or \textbf{PCSP}, aggregates early
image-to-text attention responses into sensitive and context token sets;
\textbf{Sensitive-Region State Update}, or \textbf{SRSU}, evaluates sensitive query rows while
retaining global keys and values; and \textbf{Context Anchor Refresh}, or \textbf{CAR}, periodically
recomputes all states to limit cache drift. We call this acceleration
configuration with global classifier-free guidance \textbf{\emph{SAFE-Core}}.
\textbf{Sensitivity-Weighted CFG}, or \textbf{SW-CFG}, reuses the same
sensitivity field as SAFE-DiT's prompt-alignment guidance module. It changes
prediction-level guidance rather than attention execution. This decomposition
makes the claim boundary explicit: the fast-path rewrite is exact,
\textbf{SRSU/CAR} provides controlled approximate scheduling, and
\textbf{SW-CFG} targets prompt alignment without reintroducing a spatial
self-attention mask.

Across kernel diagnostics, matched-output checks, high-resolution stress tests,
and multiple DiT-style backbones, \textbf{SAFE-DiT} improves the speed--memory frontier
and enables resolutions where dense inference exhausts memory. Fidelity uses
paired prompts and seeds, perceptual distances, and blind pairwise judgments;
automatic preference rewards provide auxiliary prompt-alignment evidence rather
than proof of perceptual superiority \cite{xu2023imagereward,wu2023hpsv2}.

Our contributions are summarized as follows:
\begin{itemize}
    \item We identify and characterize \textbf{\emph{Mask-Induced Dispatch Tax}},
    showing that a mathematically redundant mask can erase the intended
    efficiency of spatially adaptive DiT inference by changing
    attention-backend dispatch.

    \item We formulate an exact semantics criterion for \textbf{mask elision}:
    additive masks that are constant within each query row can be removed
    without changing normalized attention, while padding, causal, block, and
    non-uniform bias masks are conservatively retained.

    \item We introduce \textbf{SAFE-DiT}, which combines this exact fast-path execution
    rewrite with mask-free, prompt-conditioned query scheduling and
    anchor-based cache refresh. By reporting \textbf{SAFE-Core} and
    \textbf{SAFE-DiT+SW} side by side, the evaluation distinguishes acceleration
    fidelity from the guidance module's alignment effect.
\end{itemize}

\section{Related Work}
\label{sec:relatedwork}

\paragraph{Diffusion Transformers and Spatial Guidance.}
Modern text-to-image systems progress from denoising diffusion and classifier-free guidance to latent diffusion and Transformer-based denoisers~\cite{ho2020ddpm,dhariwal2021guided,ho2022cfg,rombach2022ldm,vaswani2017attention,dosovitskiy2021vit,peebles2023dit,bao2023uvit}, with large-scale systems relying on scalable Transformer or rectified-flow backbones whose attention cost grows quadratically with resolution~\cite{nichol2022glide,ramesh2022dalle2,saharia2022imagen,yu2022parti,balaji2022ediffi,podell2024sdxl,chen2023pixartalpha,chen2024pixartsigma,esser2024sd3,gao2024lumina}. Spatially adaptive guidance exploits a related non-uniformity---cross-attention exposes text-region correspondences and S-CFG assigns region-dependent guidance~\cite{hertz2023prompttoprompt,tang2023daam,chefer2023attendexcite,shen2024scfg}. SAFE-DiT shares this premise but asks which spatial decisions can be made without passing redundant masks into self-attention.

\paragraph{Training-free Acceleration.}
Diffusion inference is accelerated by shortening the trajectory, reusing computation, or updating selected regions. Numerical solvers cut sampling cost~\cite{song2021ddim,lu2022dpmsolver,lu2022dpmsolverpp}, and distillation reduces the step count with extra training~\cite{salimans2022progressive,song2023consistency,luo2023lcm,sauer2024ladd,li2023snapfusion}. Training-free methods instead exploit inference-time redundancy, reusing or forecasting intermediate features along the trajectory~\cite{ma2024deepcache,wimbauer2023blockcache,chen2024deltadit,selvaraju2024fora,liu2024teacache,liu2025taylorseer,cui2026dpcache}. Closest to SAFE-DiT are token- and region-aware methods that cache or reallocate computation over tokens or regions~\cite{zou2024toca,liu2025fastcache,zheng2025clusca,liu2026ras,lin2026sdit,yin2026accelaes}. These works already establish partial-query updates, cached context, dense refresh, and spatially varying guidance as useful ingredients; Appendix~\ref{sec:supp_prior_boundary} gives a component-level comparison. SAFE-DiT's distinct contribution is narrower: it identifies and certifies redundant self-attention masks as an execution-level dispatch tax, then expresses prompt-sensitive spatial scheduling outside the taxed mask interface.

\paragraph{Fast Attention and Mask-aware Execution.}
Efficient attention kernels are essential for high-resolution DiTs: FlashAttention variants provide IO-aware exact attention~\cite{dao2022flashattention,dao2023flashattention2,shah2024flashattention3}, and PyTorch SDPA selects among backends according to the supplied inputs~\cite{pytorch2026sdpa}. A separate line accelerates \emph{necessary} masked or custom attention~\cite{wang2025flashmask,dong2025flexattention}, and spatial-control methods show why masks are useful interfaces~\cite{avrahami2022blendeddiffusion,meng2022sdedit,zhang2023controlnet,li2023gligen,bar2023multidiffusion}. SAFE-DiT addresses the complementary case where the mask is \emph{not} necessary: it removes all-valid image self-attention masks, which preserve the visible key set, while retaining cross-attention text padding masks, which carry conditioning. It is thus not a new kernel but a semantics-aware rewrite that keeps spatial adaptation off the taxed self-attention interface whenever mathematically safe.

\section{Methodology}
\label{sec:method}

SAFE-DiT accelerates high-resolution Diffusion Transformers, or DiTs, by separating an operator-preserving execution rewrite from approximation-based spatial scheduling. Rather than using a spatial self-attention mask simultaneously as a semantic constraint and a routing mechanism, SAFE-DiT asks two questions: does the supplied mask change the normalized attention distribution, and, when it does not, how can spatially non-uniform computation be represented without that mask? The resulting design removes only provably redundant image self-attention masks, retains semantics-bearing text masks, and moves spatial adaptation off the mask interface entirely, building on the standard Transformer attention abstraction and on evidence that text--image cross-attention provides useful spatial attribution in diffusion models~\cite{vaswani2017attention,tang2023daam,hertz2023prompttoprompt,chefer2023attendexcite}.

\subsection{Problem Setup and Method Overview}
\label{subsec:preliminaries}

Let \(X_t\in\mathbb{R}^{N\times d}\) denote the image latent tokens at denoising step \(t\), where \(N\) is the number of image tokens and \(d\) is the hidden dimension. Let \(Y\in\mathbb{R}^{M\times d_y}\) denote the text-token representation. For one attention head, we define the projected queries, keys, and values as:
\begin{equation}
    Q=XW_Q,\qquad K=ZW_K,\qquad V=ZW_V,
    \label{eq:qkv_projection}
\end{equation}
where \(X\) supplies queries, \(Z\) supplies keys and values, and \(W_Q,W_K,W_V\) are learned projections. With an additive mask \(m\in(\mathbb{R}\cup\{-\infty\})^{n_q\times n_k}\), scaled dot-product attention is:
\begin{equation}
    \mathrm{SDPA}(Q,K,V,m)
    =
    \mathrm{Softmax}
    \left(
        \frac{QK^\top}{\sqrt{d_h}} + m
    \right)V,
    \label{eq:sdpa}
\end{equation}
where \(d_h\) is the per-head dimension. We use \(m=\varnothing\) to denote that no mask argument is supplied, equivalently a zero additive bias at the mathematical operator level.

\begin{figure*}[t]
  \centering
  \includegraphics[width=
  \linewidth]{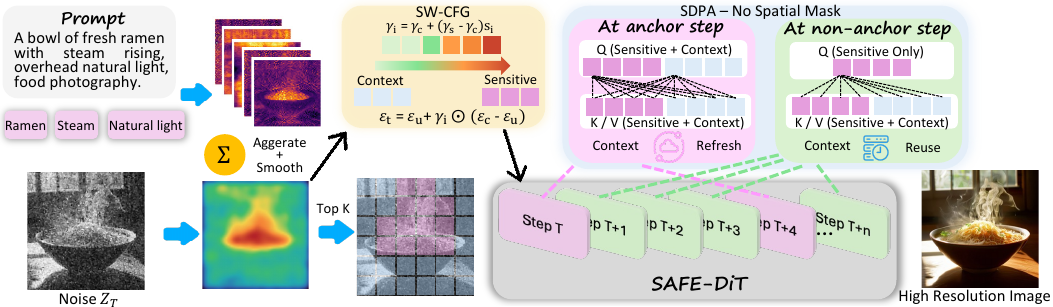}
  \caption{\textbf{SAFE-DiT pipeline.} A short warm-up aggregates image-to-text
  attention into a sensitivity map and fixes sensitive/context tokens. Anchor
  steps refresh all tokens; non-anchor steps update sensitive queries, reuse
  context keys/values, elide certified redundant self-attention masks, and keep
  text masks. SW-CFG applies sensitivity-weighted per-token guidance.}
  \label{fig:pipeline}
  \vspace{-8pt}
\end{figure*}

For clarity, we write a DiT block as image self-attention followed by image-text cross-attention:
\begin{equation}
    \mathcal{A}_{\mathrm{self}}(X_t;m_{\mathrm{img}})
    =
    \mathrm{SDPA}(Q_x,K_x,V_x,m_{\mathrm{img}}),
    \label{eq:self_attn}
\end{equation}
and the corresponding image-text cross-attention as:
\begin{equation}
    \mathcal{A}_{\mathrm{cross}}(X_t,Y;m_{\mathrm{text}})
    =
    \mathrm{SDPA}(Q_x,K_y,V_y,m_{\mathrm{text}}).
    \label{eq:cross_attn}
\end{equation}
Here \(m_{\mathrm{img}}\) is an image-token mask and \(m_{\mathrm{text}}\) is a text-token padding mask. For joint-attention backbones, Eq.~\eqref{eq:cross_attn} denotes the image-query/text-key sub-block of the joint attention matrix. SAFE-DiT classifies masks by their effect on attention, not by tensor name: an all-valid image mask may be removable, whereas a text padding mask changes valid keys and must be retained.

As shown in Fig.~\ref{fig:pipeline}, SAFE-DiT combines an exact fast-path execution rewrite, mask-free spatial scheduling, drift-control refresh, and a prompt-alignment guidance module. First, the SAFE execution rewrite exactly removes masks that induce only a row-wise constant shift of the attention logits. Second, Prompt-Conditioned Sensitivity Partitioning, or PCSP, converts early image-to-text attention responses into a sensitive set \(\mathcal{S}\) and a context set \(\mathcal{C}\). Third, Sensitive-Region State Update, or SRSU, evaluates only sensitive query and row-wise block outputs, while Context Anchor Refresh, or CAR, periodically performs a dense update to bound cache drift. Finally, Sensitivity-Weighted CFG, or SW-CFG, adjusts guidance strength across space using the same sensitivity field. We use \emph{SAFE-Core} for the acceleration-fidelity operating point with the original global CFG scale, and \emph{SAFE-DiT+SW} for the operating point that includes the SW-CFG guidance module.

\subsection{Exact Semantics-Aware Fast-path Rewrite}
\label{subsec:semantics_aware_rewrite}

Let \(L=QK^\top/\sqrt{d_h}\) be the unmasked attention logits; for query row \(i\), define the masked attention distribution as:
\begin{equation}
    \pi_i(m;L)
    =
    \mathrm{Softmax}\!\left(L_{i,:}+m_{i,:}\right),
    \label{eq:visible_set}
\end{equation}
where \(\pi_i(m;L)\) denotes the normalized attention distribution for query \(i\), and \(m_{i,:}\) is the \(i\)-th mask row. A mask is universally removable when \(\pi_i(m;L)=\pi_i(\varnothing;L)\) for every query row and arbitrary finite logits \(L\). For additive masks, this is characterized by the row-constant class:
\begin{equation}
    \mathcal{R}
    =
    \left\{
    m\;\middle|\;
    \forall i,\ \exists c_i\in\mathbb{R}
    \text{ such that } m_{ij}=c_i,\ \forall j
    \right\}.
    \label{eq:redundant_mask}
\end{equation}
For batched or head-specific masks, the criterion is applied independently to every broadcasted query row. The row-constant invariance follows from:
\begin{equation}
    \mathrm{Softmax}(L_{i,:}+c_i\mathbf{1})
    =
    \mathrm{Softmax}(L_{i,:}),
    \label{eq:row_constant_invariance}
\end{equation}
where \(\mathbf{1}\) is the all-ones row vector; the common factor \(\exp(c_i)\) cancels in row-wise normalization. Conversely, a non-constant row changes the attention distribution for some logits. Thus Eq.~\eqref{eq:redundant_mask} is sufficient and, for equality under arbitrary logits, necessary. An all-valid Boolean mask and an all-zero additive mask are special cases. A finite non-uniform bias is not removable even if every key remains visible, and any mask containing blocked positions must also be preserved.

SAFE-DiT therefore applies the certified rewrite:
\begin{equation}
\begin{aligned}
    &\mathrm{SAFE\text{-}SDPA}(Q,K,V,m)\\[-0.2em]
    &\quad =
    \begin{cases}
    \mathrm{SDPA}(Q,K,V,\varnothing),
    & m\in\mathcal{R},\\
    \mathrm{SDPA}(Q,K,V,m),
    & m\notin\mathcal{R},
    \end{cases}
\end{aligned}
    \label{eq:safe_sdpa}
\end{equation}
where \(\mathcal{R}\) is the row-constant mask class in Eq.~\eqref{eq:redundant_mask}, and \(m=\varnothing\) denotes a mask-free SDPA call as in Eq.~\eqref{eq:sdpa}. The implementation does not scan the mask at runtime; it applies the rewrite only when mask provenance certifies an all-valid, zero, or row-broadcast constant mask, and otherwise keeps the original call, so redundant image self-attention masks are removed while \(m_{\mathrm{text}}\), causal, and block masks are retained. Equation~\eqref{eq:safe_sdpa} is exact in real arithmetic; fast and masked kernels may still differ by small accumulation-order effects, so we evaluate operator equivalence and bitwise identity separately.

\subsection{Prompt-Conditioned Sensitivity Partitioning}
\label{subsec:pcsp}

PCSP estimates where the prompt warrants more frequent computation using early image-to-text attention, following prior evidence that diffusion cross-attention maps localize prompt concepts and support editing or semantic guidance~\cite{hertz2023prompttoprompt,tang2023daam,chefer2023attendexcite}. We use the following image-to-text attention tensor:
\begin{equation}
    A_{t,\ell,h}\in[0,1]^{N\times M},
    \label{eq:cross_attention_tensor}
\end{equation}
where \(A_{t,\ell,h}\) denotes the attention probabilities from image queries to text keys at denoising step \(t\), layer \(\ell\), and head \(h\). In a joint-attention backbone, \(A_{t,\ell,h}\) is the corresponding image-to-text probability submatrix after the model's native masking and normalization. For image token \(i\), we aggregate responses to content tokens:
\begin{equation}
    \tilde{s}_i
    =
    \frac{1}{|\mathcal{W}||\mathcal{L}||\mathcal{H}|}
    \sum_{t\in\mathcal{W}}
    \sum_{\ell\in\mathcal{L}}
    \sum_{h\in\mathcal{H}}
    \sum_{j\in\mathcal{T}}
    w_j A_{t,\ell,h}[i,j],
    \label{eq:raw_sensitivity}
\end{equation}
where \(\mathcal{W}\) is a short dense warm-up window, \(\mathcal{L}\) and \(\mathcal{H}\) are fixed layer and head subsets, \(\mathcal{T}\) contains non-padding content-token indices, and \(w_j\geq0\) are normalized content weights with \(\sum_{j\in\mathcal{T}}w_j=1\). Padding and special tokens receive zero weight. Instrumentation runs only during the warm-up calls and retains a single \(N\)-dimensional score vector rather than full attention tensors.

We smooth and normalize the score on the latent grid:
\begin{equation}
    s
    =
    \mathrm{Norm}
    \left(
        G_\sigma * \tilde{s}
    \right),
    \label{eq:normalized_sensitivity}
\end{equation}
where \(G_\sigma\) is a Gaussian kernel and \(\mathrm{Norm}(\cdot)\) maps the field to \([0,1]\). Given a target sensitive-token ratio \(\rho\), PCSP selects exactly \(k=\lceil\rho N\rceil\) tokens:
\begin{equation}
    \mathcal{S}
    =
    \mathrm{TopK}(s,k),
    \qquad
    \mathcal{C}
    =
    \{1,\ldots,N\}\setminus\mathcal{S},
    \label{eq:partition}
\end{equation}
where \(\mathrm{TopK}(s,k)\) returns the \(k\) image-token indices with the largest normalized sensitivity scores, and \(\mathcal{C}\) is the complementary context set. The partition is fixed after warm-up, avoiding repeated routing overhead, and requires neither an external segmentation model nor a regional self-attention mask. It only supplies indices to the scheduler.

\subsection{Fast-path Sensitivity-Conditioned State Scheduling}
\label{subsec:state_scheduling}

Let \(H_{t,\ell}\in\mathbb{R}^{N\times d_\ell}\) be the hidden states at denoising step \(t\) and layer \(\ell\). Row-selection operators \(P_{\mathcal{S}}\) and \(P_{\mathcal{C}}\) define selected states as:
\begin{equation}
    H_{t,\ell}^{\mathcal{S}} = P_{\mathcal{S}}H_{t,\ell},
    \qquad
    H_{t,\ell}^{\mathcal{C}} = P_{\mathcal{C}}H_{t,\ell},
    \label{eq:state_split}
\end{equation}
where \(H_{t,\ell}^{\mathcal{S}}\) and \(H_{t,\ell}^{\mathcal{C}}\) are the sensitive and context rows of \(H_{t,\ell}\), respectively. Let \(\mathcal{A}\subseteq\{1,\ldots,T\}\) be the anchor-step set and let \(a(t)=\max\{\tau\in\mathcal{A}:\tau<t\}\) denote the most recent anchor before a non-anchor step \(t\). The context cache stores layerwise context states from that anchor:
\begin{equation}
    \widetilde{H}_{t,\ell}^{\mathcal{C}}
    =
    P_{\mathcal{C}}H_{a(t),\ell},
\label{eq:context_cache}
\end{equation}
where \(\widetilde{H}_{t,\ell}^{\mathcal{C}}\) is the cached context state reused at non-anchor step \(t\).
At a non-anchor step, SRSU evaluates a row-selective version \(B_\ell^{\mathcal{S}}\) of the \(\ell\)-th DiT block:
\begin{equation}
    H_{t,\ell+1}^{\mathcal{S}}
    =
    B_\ell^{\mathcal{S}}
    \left(
        H_{t,\ell}^{\mathcal{S}},
        \widetilde{H}_{t,\ell}^{\mathcal{C}},
        Y
    \right),
    \qquad
    H_{t,\ell+1}^{\mathcal{C}}
    =
    \widetilde{H}_{t,\ell+1}^{\mathcal{C}},
    \label{eq:srsu_update}
\end{equation}
where \(B_\ell^{\mathcal{S}}\) preserves the original block order and parameters but computes attention outputs, residual updates, and row-wise MLP outputs only for \(\mathcal{S}\). Context output rows are copied from the cache.

For self-attention, sensitive queries retain global spatial context:
\begin{equation}
    O_{\mathrm{self}}^{\mathcal{S}}
    =
    \mathrm{SDPA}
    \left(
        Q_{\mathcal{S}},
        K_{\mathcal{S}\cup\mathcal{C}},
        V_{\mathcal{S}\cup\mathcal{C}},
        \varnothing
    \right),
    \label{eq:query_selective_self_attention}
\end{equation}
where \(Q_{\mathcal{S}}\) is formed from current sensitive states, while \(K_{\mathcal{S}\cup\mathcal{C}}\) and \(V_{\mathcal{S}\cup\mathcal{C}}\) are assembled in the original token order from current sensitive states and cached context states. SRSU therefore restricts query rows, not visible keys: each updated token can still attend to the full image context. For cross-attention, the row-selective output is:
\begin{equation}
    O_{\mathrm{cross}}^{\mathcal{S}}
    =
    \mathrm{SDPA}
    \left(
        Q_{\mathcal{S}},
        K_y,
        V_y,
        m_{\mathrm{text}}
    \right),
    \label{eq:query_selective_cross_attention}
\end{equation}
where \(K_y\), \(V_y\), and \(m_{\mathrm{text}}\) are the text keys, values, and padding mask from Eq.~\eqref{eq:cross_attn}. Thus, the same query reduction is used without removing text semantics. Relative to dense self-attention, the dominant attention term changes from \(O(N^2d_h)\) to \(O(|\mathcal{S}|Nd_h)\), while row-wise block computation scales from \(N\) to \(|\mathcal{S}|\), up to the cost of forming global keys and values.

CAR bounds the approximation error introduced by stale context states. For every \(t\in\mathcal{A}\), SAFE-DiT evaluates the original dense block:
\begin{equation}
    H_{t,\ell+1}=B_\ell(H_{t,\ell},Y),
    \label{eq:anchor_dense_update}
\end{equation}
and refreshes the layerwise context cache:
\begin{equation}
    \widetilde{H}_{t,\ell+1}^{\mathcal{C}}
    \leftarrow
    P_{\mathcal{C}}H_{t,\ell+1},
    \label{eq:cache_refresh}
\end{equation}
where the arrow denotes replacement of the cached context rows after a dense anchor update. For \(t\notin\mathcal{A}\), Eq.~\eqref{eq:srsu_update} is used. The ratio \(\rho\) and anchor schedule \(\mathcal{A}\) expose the principal speed--fidelity trade-off: smaller \(\rho\) and fewer anchors reduce computation, whereas larger values track the dense trajectory more closely.

\subsection{Sensitivity-Weighted Guidance and Inference}
\label{subsec:swcfg_algorithm}

SW-CFG reuses the sensitivity field as a prompt-alignment guidance module. Let \(f_c(X_t,Y)_i\) and \(f_u(X_t)_i\) be conditional and unconditional predictions for token \(i\). Standard CFG uses a global scale \(\gamma\). SW-CFG instead defines:
\begin{equation}
    \gamma_i
    =
    \gamma_{\mathrm{ctx}}
    +
    \left(
        \gamma_{\mathrm{sens}}-\gamma_{\mathrm{ctx}}
    \right)s_i,
    \label{eq:gamma_continuous}
\end{equation}
where \(s_i\) is the normalized sensitivity score from Eq.~\eqref{eq:normalized_sensitivity}, and \(\gamma_{\mathrm{ctx}}\) and \(\gamma_{\mathrm{sens}}\) are the context and sensitive-region guidance scales. The guided token prediction is:
\begin{equation}
    \hat{f}_i
    =
    f_u(X_t)_i
    +
    \gamma_i
    \left(
        f_c(X_t,Y)_i - f_u(X_t)_i
    \right),
    \label{eq:swcfg}
\end{equation}
where \(\hat{f}_i\) denotes the SW-CFG prediction for token \(i\). A binary variant assigns \(\gamma_{\mathrm{sens}}\) to \(\mathcal{S}\) and \(\gamma_{\mathrm{ctx}}\) to \(\mathcal{C}\). Because this operation is applied after the conditional and unconditional forward passes, it introduces no self-attention mask. Setting \(\gamma_{\mathrm{sens}}=\gamma_{\mathrm{ctx}}=\gamma\) recovers SAFE-Core and preserves the dense model's original guidance rule; using unequal scales gives the SAFE-DiT+SW operating point.

At inference, dense warm-up steps build the partition and context cache; anchor steps then refresh all states while non-anchor steps update only sensitive rows, reuse cached context rows, elide certified-redundant self-attention masks, and retain text masks, with the sampler applying global CFG for SAFE-Core or SW-CFG for SAFE-DiT+SW. Any mask the rewrite cannot certify as redundant is left untouched, so correctness never depends on the elision firing. This makes the method's claims explicit: mask elision is an exact execution rewrite, SRSU and CAR give a controlled speed--fidelity trade-off, and SW-CFG sets the prompt-alignment operating point.

\section{Experiments and Results}
\label{sec:experiments}

\begin{table*}[!t]
\centering
\caption{
Main Lumina-Next comparison on DrawBench using 200 prompts and three paired
seeds per prompt. Deltas are computed relative to Dense; LPIPS and pixel
difference are paired distances to Dense. Dense+FP applies only the certified
fast-path rewrite, SAFE-Core additionally applies mask-free spatial scheduling
with global CFG, and SAFE-DiT+SW further enables SW-CFG. Bold values indicate
the best result in each column among non-Dense methods, with ties retained.
}
\label{tab:baseline_comparison}
\scriptsize
\setlength{\tabcolsep}{3pt}
\renewcommand{\arraystretch}{0.92}
\resizebox{0.9\textwidth}{!}{%
\begin{tabular}{l
>{\columncolor{effbg}}c
>{\columncolor{effbg}}c
>{\columncolor{effbg}}c
>{\columncolor{rewbg}}c
>{\columncolor{rewbg}}c
>{\columncolor{rewbg}}c
>{\columncolor{rewbg}}c
>{\columncolor{fidbg}}c
>{\columncolor{fidbg}}c}
\toprule
\multicolumn{1}{l}{\textbf{Method}} &
\multicolumn{3}{c}{\cellcolor{effhead}\textbf{Efficiency}} &
\multicolumn{4}{c}{\cellcolor{rewhead}\textbf{Reward Metrics}} &
\multicolumn{2}{c}{\cellcolor{fidhead}\textbf{Fidelity to Dense}} \\
\cmidrule(lr){2-4}
\cmidrule(lr){5-8}
\cmidrule(l){9-10}
Method &
Lat., s \(\downarrow\) &
Speedup \(\uparrow\) &
Mem., GB \(\downarrow\) &
IR \(\uparrow\) &
HPSv2 \(\uparrow\) &
\(\Delta\)IR &
\(\Delta\)HPS &
LPIPS \(\downarrow\) &
Pixel diff. \(\downarrow\) \\
\midrule

\rowcolor{densebg}
Dense
& 8.983
& \(1.00\times\)
& 11.34
& 0.477
& 0.2787
& 0.000
& 0.0000
& --
& -- \\

\midrule

FastCache~\cite{liu2025fastcache}
& \textbf{2.559}
& \(\mathbf{3.51\times}\)
& 12.65
& -0.573
& 0.2555
& -1.050
& -0.0232
& 0.484
& 18.80 \\

DPCache~\cite{cui2026dpcache}
& 3.930
& \(2.29\times\)
& 11.71
& 0.494
& 0.2794
& +0.017
& +0.0007
& 0.105
& 6.71 \\

TaylorSeer~\cite{liu2025taylorseer}
& 4.300
& \(2.09\times\)
& \textbf{11.34}
& 0.419
& 0.2778
& -0.058
& -0.0009
& 0.116
& 7.65 \\

AccelAes~\cite{yin2026accelaes}
& 4.319
& \(2.08\times\)
& 11.96
& \textbf{0.521}
& 0.2805
& \textbf{+0.044}
& +0.0018
& 0.092
& 6.08 \\

ClusCa~\cite{zheng2025clusca}
& 5.413
& \(1.66\times\)
& 12.53
& 0.384
& 0.2747
& -0.093
& -0.0040
& 0.187
& 8.69 \\

SDiT~\cite{lin2026sdit}
& 5.521
& \(1.63\times\)
& 11.63
& 0.243
& 0.2719
& -0.234
& -0.0068
& 0.248
& 12.59 \\

RAS~\cite{liu2026ras}
& 5.999
& \(1.50\times\)
& 11.76
& 0.475
& 0.2787
& -0.001
& -0.0001
& 0.016
& 1.91 \\

\(\Delta\)-DiT~\cite{chen2024deltadit}
& 6.077
& \(1.48\times\)
& 12.36
& 0.274
& 0.2727
& -0.202
& -0.0061
& 0.239
& 13.23 \\

TeaCache~\cite{liu2024teacache}
& 8.689
& \(1.03\times\)
& \textbf{11.34}
& 0.469
& 0.2784
& -0.008
& -0.0003
& \textbf{0.014}
& \textbf{1.34} \\

FORA~\cite{selvaraju2024fora}
& 9.189
& \(0.98\times\)
& 12.39
& 0.353
& 0.2708
& -0.123
& -0.0079
& 0.439
& 19.32 \\

S-CFG~\cite{shen2024scfg}
& 9.210
& \(0.98\times\)
& 11.68
& 0.443
& 0.2779
& -0.034
& -0.0008
& 0.125
& 8.45 \\

\midrule

Dense+FP
& 5.442
& \(1.65\times\)
& \textbf{11.34}
& 0.478
& 0.2787
& +0.001
& +0.0000
& 0.040
& 3.43 \\

SAFE-Core
& 3.339
& \(2.69\times\)
& 11.68
& 0.486
& 0.2795
& +0.009
& +0.0008
& 0.043
& 3.61 \\

\rowcolor{safebg}
\textbf{SAFE-DiT+SW}
& 3.344
& \(2.69\times\)
& 11.69
& 0.520
& \textbf{0.2807}
& +0.043
& \textbf{+0.0020}
& 0.119
& 7.15 \\

\bottomrule
\end{tabular}
}
\end{table*}

\subsection{Experimental Setup}
\label{subsec:exp_setup}

\paragraph{Backbones and Prompts.}
Lumina-Next is the primary backbone because its high-resolution inference exposes a strong attention bottleneck and represents recent scalable DiT-style systems~\cite{gao2024lumina}. We additionally evaluate SD3-Medium~\cite{esser2024sd3}, FLUX.1-dev~\cite{blackforestlabs2024flux}, and PixArt-\(\Sigma\)-style high-resolution settings~\cite{chen2024pixartsigma} to test whether the conclusions depend on a single implementation stack. The main benchmark uses 200 DrawBench prompts~\cite{saharia2022imagen} with three paired seeds per prompt, giving 600 generations per method. We also use a 300-prompt stratified suite informed by MS-COCO, PartiPrompts, GenEval, and T2I-CompBench prompt categories~\cite{lin2014coco,yu2022parti,ghosh2023geneval,huang2023t2icompbench}, a high-resolution stress suite from \(1024^2\) to \(3072^2\), a blinded human study, and a 100-prompt VLM subset as an auxiliary perceptual check.

\paragraph{Protocol and Metrics.}
All methods are evaluated with paired prompts, paired seeds, the same sampler, the same number of function evaluations, and bf16 precision unless otherwise stated. Latency is measured after three warm-up runs with CUDA synchronization before and after each timed run. We record both wall-clock latency and peak GPU memory under the same inference configuration. ImageReward, abbreviated IR, is the primary prompt-aware reward metric~\cite{xu2023imagereward}, with HPSv2~\cite{wu2023hpsv2}, LPIPS~\cite{zhang2018lpips}, pixel difference, human preference, and auxiliary VLM win/tie/loss judgments as complementary quality and fidelity checks. SDPA backend dispatch is logged, since it varies with the supplied inputs across FlashAttention-style, memory-efficient, and math kernels~\cite{dao2022flashattention,dao2023flashattention2,pytorch2026sdpa}.

\paragraph{Systems Protocol.}
All measurements use one NVIDIA RTX PRO 6000 Blackwell GPU under PyTorch 2.8 / CUDA 12.8 with the Flash, memory-efficient, and math SDPA backends enabled. Reported end-to-end latency covers prompt encoding, denoising, PCSP, VAE decoding, and post-processing after three warm-up runs with CUDA synchronization; denoiser-only latency is recorded separately. With VAE scale factor 8 and patch size 2, \(1024^2\)/\(2560^2\)/\(3072^2\) map to 4{,}096/25{,}600/36{,}864 latent tokens. Appendix~\ref{sec:supp_protocol} gives full implementation and timing details.

\subsection{Main Results}
\label{subsec:main_results}

Certified removable masks agree with an explicit FP32 reference, \(\mathrm{softmax}(QK^\top/\sqrt{d_h}+m)V\), to within \(9\times10^{-8}\). The observed 0.22\% bf16 attention-output difference after removing the all-valid image self-attention mask comes from fused-kernel accumulation order, while dropping the text padding mask is not equivalent, with LPIPS 0.169. SAFE-DiT therefore elides only redundant image self-attention masks and retains text masks. The redundant mask is expensive: at head dimension 72, masked SDPA is \(4.1\)--\(5.8\times\) slower than the mask-free call, and compiled FlexAttention closes only part of the gap. The official \texttt{diffusers} Lumina processor passes the all-valid mask unconditionally, and RAS~\cite{liu2026ras} inherits this path; SAFE-DiT removes the certified-redundant mask automatically, without custom kernels. Full dispatch and exactness results are in Appendices~\ref{sec:supp_midt} and~\ref{sec:supp_exactness}.

\begin{table*}[!t]
\centering
\caption{Fair fast-path audit on high-resolution Lumina-Next. \(2560^2\) values are medians over three stress prompts; \(3072^2\) tests runnability on one prompt. ``Mask'' marks whether the all-valid image self-attention mask is kept or elided.}
\label{tab:fair_fastpath}
\scriptsize
\setlength{\tabcolsep}{3.0pt}
\renewcommand{\arraystretch}{0.88}
\resizebox{\textwidth}{!}{%
\begin{tabular}{l
>{\columncolor{effbg}}c>{\columncolor{effbg}}c>{\columncolor{effbg}}c>{\columncolor{effbg}}c>{\columncolor{effbg}}c
>{\columncolor{rewbg}}c>{\columncolor{rewbg}}c>{\columncolor{rewbg}}c
>{\columncolor{fidbg}}c}
\toprule
\multicolumn{1}{l}{\textbf{Variant}} &
\multicolumn{5}{c}{\cellcolor{effhead}\textbf{Components}} &
\multicolumn{3}{c}{\cellcolor{rewhead}\(\mathbf{2560^2}\) \textbf{Stress}} &
\multicolumn{1}{c}{\cellcolor{fidhead}\(\mathbf{3072^2}\) \textbf{Status}} \\
\cmidrule(lr){2-6}\cmidrule(lr){7-9}\cmidrule(l){10-10}
Variant & Mask & FP & PCSP/Q & FFN/cache & SW &
Lat., s \(\downarrow\) & Mem., GB \(\downarrow\) & Speedup \(\uparrow\) &
Lat./Mem. or OOM \\
\midrule
\rowcolor{densebg}
Dense-original & kept & no & -- & -- & -- & 209.3 & 94.1 & \(1.00\times\) & OOM \\
Dense + mask elision & elided & yes & -- & -- & -- & 62.6 & \textbf{24.7} & \(3.34\times\) & 126.7s / 26.9GB \\
Spatial sched. w/o elision & kept & no & on & on & on & OOM & OOM & -- & OOM \\
\midrule
SAFE query-only & elided & yes & query & -- & -- & 63.5 & 27.7 & \(3.30\times\) & 127.1s / 31.1GB \\
SAFE-Core & elided & yes & on & on & -- & 41.6 & 27.9 & \(5.03\times\) & 87.6s / 36.2GB \\
\rowcolor{safebg}
\textbf{SAFE-DiT+SW} & elided & yes & on & on & on & \textbf{41.1} & 27.9 & \(\mathbf{5.09\times}\) & \textbf{87.6s / 36.2GB} \\
\bottomrule
\end{tabular}
}
\end{table*}

As shown in Table~\ref{tab:baseline_comparison}, a paired, prompt-clustered bootstrap using 10{,}000 resamples gives significant reward gains over Dense for SAFE-DiT, \(\Delta\text{IR}=+0.043\) [\(+0.022,+0.065\)], and AccelAes, \(\Delta\text{IR}=+0.044\) [\(+0.026,+0.062\)], whereas DPCache's \(+0.017\) is not significant. Under our Lumina-Next adaptation and threshold range, FastCache is the only faster method but has no fast high-reward operating point, so the high-reward comparison is AccelAes. SAFE-DiT reaches statistically indistinguishable quality from AccelAes, with \(\Delta\)IR \(=-0.001\) and CI \([-0.016,+0.014]\), while giving a \(1.29\times\) speedup, 3.34 vs.\ 4.32\,s, as the MIDT rewrite removes a dispatch cost AccelAes still pays. The ordering holds on a 300-prompt stratified suite, where IR is 0.594 vs.\ 0.550 for DPCache.

As shown in Table~\ref{tab:fair_fastpath}, the exact rewrite is separated from the scheduling. ``Dense\,+\,mask elision'' applies only the rewrite and already removes most of the tax, reducing latency from \(209\) to \(63\)\,s and memory from \(94\) to \(25\)\,GB at \(2560^2\). SAFE-Core then adds a further \(1.5\times\) \emph{on top of} this strengthened baseline, so the scheduling gain is genuine rather than an artifact of a taxed reference. Here mask elision dominates as the tax grows with sequence length, while query scheduling is latency-neutral and the SW-CFG module adds no measurable compute.

\begin{figure*}[!t]
  \centering
  \includegraphics[width=0.92\textwidth]{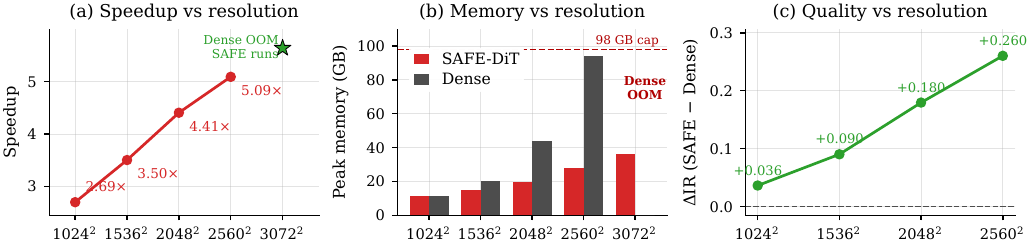}
  \caption{High-resolution speed--memory frontier. SAFE-DiT's advantage increases with resolution on Lumina-Next and enables \(3072^2\) generation where dense inference runs out of memory.}
  \label{fig:highres_frontier}
  \vspace{-4pt}
\end{figure*}

As shown in Fig.~\ref{fig:highres_frontier}, the advantage widens with resolution, from \(2.7\times\) at \(1024^2\) to \(5.1\times\) at \(2560^2\). Temporal caches materialize full attention per step. They do not reduce peak memory and run out of memory at \(2560^2\); SAFE-DiT completes \(2560^2\) and \(3072^2\) by shrinking per-step token computation. Appendix~\ref{sec:supp_frontier} reports the full high-resolution frontier.

\begin{table}[t]
\centering
\caption{Cross-backbone generalization. Results are paired and use \(1024^2\) unless noted. Speedup and \(\Delta\)IR are measured vs.\ Dense; ``Mem.'' reports peak GB from Dense to SAFE. ``Self-attn'' marks whether the backbone triggers MIDT.}
\label{tab:crossbackbone}
\footnotesize
\setlength{\tabcolsep}{4pt}
\renewcommand{\arraystretch}{0.95}
\begin{tabular}{ll
>{\columncolor{effbg}}c>{\columncolor{effbg}}c
>{\columncolor{fidbg}}c>{\columncolor{fidbg}}c}
\toprule
\multicolumn{2}{l}{\textbf{Backbone}} &
\multicolumn{2}{c}{\cellcolor{effhead}\textbf{Efficiency}} &
\multicolumn{2}{c}{\cellcolor{fidhead}\textbf{Quality}} \\
\cmidrule(lr){3-4}\cmidrule(l){5-6}
Backbone & Self-attn & Speedup \(\uparrow\) & Mem.\ \(\downarrow\) & \(\Delta\)IR & LPIPS \(\downarrow\) \\
\midrule
\rowcolor{safebg}
Lumina-Next & masked & \textbf{2.69\(\times\)} & 11.3\(\to\)11.7 & +0.043 & 0.119 \\
SD3-Medium & mask-free & 1.30\(\times\) & 18.1\(\to\)18.1 & +0.004 & \textbf{0.026} \\
FLUX.1-dev & mask-free & \textbf{5.14\(\times\)} & 26.5\(\to\)26.5 & \(-\)0.009 & 0.154 \\
PixArt-\(\Sigma\)\textsuperscript{*} & mask-free & 1.87\(\times\) & 23.2\(\to\)23.2 & +0.028 & 0.307 \\
\bottomrule
\end{tabular}
\\[2pt]
{\scriptsize \textsuperscript{*}PixArt-\(\Sigma\) at \(2048^2\); other rows at \(1024^2\).}
\end{table}

As shown in Table~\ref{tab:crossbackbone}, the cross-backbone results double as an end-to-end test of the MIDT boundary: speedup tracks the mask column. On the masked Lumina-Next backbone, the rewrite and scheduling compound, and the gain \emph{scales with resolution}, from \(2.7\) to \(5.1\times\), as the mask tax grows. Mask-free backbones such as SD3, FLUX, and PixArt cannot pay that tax, so their gain is scheduling-only and model-dependent, from \(1.3\times\) on SD3 to \(5.14\times\) on the step-heavy FLUX. Quality remains stable on Lumina-Next, SD3, FLUX, and the selected PixArt-\(\Sigma\) \(2048^2\) operating point; PixArt-\(\Sigma\) shows larger reward drops at \(1024^2\) and \(3072^2\) under the tested scheduling configuration. Appendix~\ref{sec:supp_xbackbone} expands the backbone table.

\subsection{Ablation Results}
\label{subsec:ablation}

\begin{table}[t]
\centering
\caption{Component ablation on DrawBench 200\(\times\)3 at \(1024^2\). Each row removes one module from Full-SAFE. LPIPS is drift vs.\ Dense.}
\label{tab:ablation}
\small
\setlength{\tabcolsep}{6pt}
\renewcommand{\arraystretch}{1.05}
\begin{tabular}{l>{\columncolor{effbg}}c>{\columncolor{fidbg}}c>{\columncolor{fidbg}}c}
\toprule
Variant & Lat., s \(\downarrow\) & IR \(\uparrow\) & LPIPS \(\downarrow\) \\
\midrule
\rowcolor{safebg}
Full-SAFE & 3.33 & 0.520 & 0.119 \\
w/o fast-path & 4.10 & 0.525 & 0.119 \\
w/o SRSU & 5.16 & 0.543 & 0.116 \\
w/o CAR & \textbf{2.94} & 0.471 & 0.292 \\
w/o PCSP & 3.33 & 0.492 & 0.066 \\
w/o SW-CFG & 3.34 & 0.486 & 0.043 \\
\bottomrule
\end{tabular}
\end{table}

\begin{figure}[!t]
  \centering
  \includegraphics[width=\linewidth]{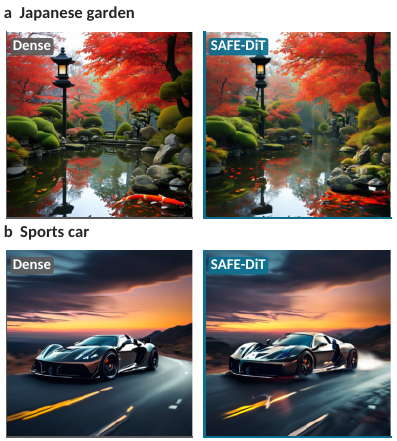}
  \caption{Paired \(2048^2\) examples with matched prompt/seed and PixArt-\(\Sigma\)-style generation. SAFE-DiT improves local detail while keeping composition close to Dense.}
  \label{fig:qualitative_generation}
  \vspace{-12pt}
\end{figure}

As shown in Table~\ref{tab:ablation}, component ablations give each module a distinct, measurable role. Removing the fast-path rewrite or SRSU increases latency while leaving paired fidelity nearly unchanged; the modest reward increase, with IR \(0.525\) and \(0.543\), respectively, versus \(0.520\) for Full-SAFE, indicates that these variants occupy slower, more conservative operating points rather than dominating the full configuration. Removing CAR is faster but drifts badly, with LPIPS rising from \(0.12\) to \(0.29\), while removing PCSP or SW-CFG leaves latency unchanged but lowers IR, separating drift control from prompt-alignment reward. The latency attribution is resolution dependent: at \(1024^2\), SRSU contributes the larger incremental latency reduction; at \(2560^2\), mask elision dominates because MIDT scales with sequence length, as shown in Table~\ref{tab:fair_fastpath}. Additional sweeps and drift curves are in Appendix~\ref{sec:supp_ablation}.

On T2I-CompBench, using 300 validation prompts per category, SAFE-Core matches Dense alignment across all categories while the SW-CFG module lifts the attribute-binding average from \(0.404\) to \(0.419\), showing that SAFE-DiT's guidance module mainly affects prompt alignment.

\subsection{Visualization}
\label{subsec:visualization}

As shown in Fig.~\ref{fig:qualitative_generation}, two paired \(2048^2\) examples improve local detail while staying compositionally close to Dense. They are illustrative; paired metrics and a blinded human study support the preservation claim. SAFE-Core has no prompt-level loss against Dense+FP on visual quality and passes the 5\% non-inferiority criterion. VLM pairwise checks are auxiliary: GPT-5 scores SAFE-DiT vs.\ Dense at 37/32/31 win/tie/loss, while GPT-4o judges most pairs tied. Failures concentrate in text rendering and dense multi-object scenes. Appendix~\ref{sec:supp_human} details the human-study protocol.

\section{Conclusion}
\label{sec:conclusion}

SAFE-DiT reframes spatial adaptation for high-resolution Diffusion Transformers as semantics-aware execution. It removes only masks that induce row-wise constant shifts, preserves text padding masks, and schedules spatial computation through PCSP, SRSU, CAR, and SW-CFG. Across MIDT diagnostics, equivalence checks, baseline comparisons, ablations, high-resolution stress tests, and blind preference studies, it improves the speed--memory frontier while preserving perceptual quality under matched prompts and seeds. By separating exact fast-path rewriting from approximate scheduling and the SW-CFG guidance module, SAFE-DiT keeps acceleration fidelity and prompt-alignment effects inspectable rather than entangled. Its gains are largest where redundant masked self-attention is present, and on hard cases such as text rendering, dense multi-object scenes, and precise layout, SAFE-DiT tracks rather than reshapes the backbone. These results support a narrow principle: spatial adaptation should use semantics-preserving scheduling rather than attention masks.

\FloatBarrier
{
    \small
    \bibliographystyle{ieeenat_fullname}
    \bibliography{main}
}

\clearpage
\appendix
\section*{Appendix Overview}

This appendix provides the full experimental record behind the main paper:
the component-level relation to the closest spatial accelerators
(Sec.~\ref{sec:supp_prior_boundary}); the implementation and timing protocol
(Sec.~\ref{sec:supp_protocol}); the kernel-level and third-party evidence for
the Mask-Induced Dispatch Tax (Sec.~\ref{sec:supp_midt}); the complete
mask-exactness matrix (Sec.~\ref{sec:supp_exactness}); paired significance for
every comparison method and the FastCache threshold sweep
(Sec.~\ref{sec:supp_stats}); the fair fast-path decomposition and instrumented
breakdown (Sec.~\ref{sec:supp_decomp}); the full high-resolution frontier with
acceleration baselines (Sec.~\ref{sec:supp_frontier}); the A800
cross-architecture evaluation (Sec.~\ref{sec:supp_a800}); the complete
cross-backbone table (Sec.~\ref{sec:supp_xbackbone}); module ablations,
hyperparameter sweeps, and the CAR drift curve (Sec.~\ref{sec:supp_ablation});
the perceptual evaluation details (Sec.~\ref{sec:supp_perceptual}); and the
completed human-study protocol with its non-inferiority analysis
(Sec.~\ref{sec:supp_human}).

\FloatBarrier

\section{Component-Level Relation To Prior Work}
\label{sec:supp_prior_boundary}

The closest spatial accelerators share several high-level ingredients with
SAFE-DiT, so we state the novelty boundary at component granularity rather than
using only a broad ``complementary'' distinction. As shown in
Table~\ref{tab:supp_prior_boundary}, the comparison is a boundary map, not a
novelty count: partial-query updates, cached context, dense refresh, and spatial
guidance are prior ingredients. The contribution claimed here is narrower:
certified exact rewriting of redundant image self-attention masks, plus
prompt-sensitive scheduling that keeps spatial adaptation off the self-attention
mask interface.

\begin{table*}[!b]
\centering
\caption{Component-level comparison with the closest spatial/token accelerators.
Shared checkmarks mark prior ingredients, not novelty claims. SAFE-DiT's
distinct claim is the certified exact rewrite that removes redundant image
self-attention masks before dispatch, together with a mask-free scheduling
decomposition.}
\label{tab:supp_prior_boundary}
\scriptsize
\setlength{\tabcolsep}{4pt}
\renewcommand{\arraystretch}{1.05}
\resizebox{\textwidth}{!}{%
\begin{tabular}{@{}ll>{\columncolor{effbg}}c>{\columncolor{effbg}}c>{\columncolor{effbg}}c>{\columncolor{effbg}}c>{\columncolor{rewbg}}c>{\columncolor{fidbg}}c@{}}
\toprule
Method & Region signal & Partial Q & Global K/V & Context cache &
Dense refresh & Spatial CFG & Certified exact rewrite \\
\midrule
RAS & model focus & yes & yes & yes & yes & no & no \\
SDiT & semantic region & yes & yes & yes & yes & no & no \\
AccelAes & cross-attn/aesthetic & yes & yes & yes & yes & yes & no \\
\rowcolor{safebg}
SAFE-DiT & prompt sensitivity & yes & yes & yes & yes & yes & yes \\
\bottomrule
\end{tabular}
}
\end{table*}

\section{Implementation And Timing Protocol}
\label{sec:supp_protocol}

All measurements use one NVIDIA RTX PRO 6000 Blackwell Server Edition GPU
(97{,}887\,MiB, driver 590.44.01) with Python 3.12.3, CUDA 12.8,
PyTorch 2.8.0+cu128, Triton 3.4.0, diffusers 0.30.3, and transformers 4.44.2.
The PyTorch SDPA flash, memory-efficient, and math backends are enabled;
standalone \texttt{flash-attn} and xFormers are not installed. The generation
path uses no \texttt{torch.compile}; compiled FlexAttention appears only in the
attention microbenchmark. Unless stated otherwise the prompt batch size is one,
classifier-free guidance is packed as a two-sample transformer batch,
Lumina-Next uses 30 bf16 denoising steps, and the text encoder uses a maximum
prompt length of 24 tokens. With VAE scale factor 8 and patch size 2,
\(1024^2\), \(2560^2\), and \(3072^2\) correspond to 4{,}096, 25{,}600, and
36{,}864 latent image tokens.

Microbenchmarks use 100 warm-up and 500 to 1000 timed repetitions; end-to-end
runs use at least three warm-up and 30 timed repetitions with CUDA
synchronization before and after each measurement. End-to-end latency includes
prompt encoding, denoising, PCSP construction when enabled, VAE decoding, and
post-processing; denoiser-only latency is the synchronized sum of
\texttt{transformer.forward} calls. Peak memory is read after
\texttt{reset\_peak\_memory\_stats}. OOM is defined under a fixed memory cap and
allocator setting.

\section{Mask-Induced Dispatch Tax}
\label{sec:supp_midt}

\begin{figure}[!b]
  \centering
  \includegraphics[width=\linewidth]{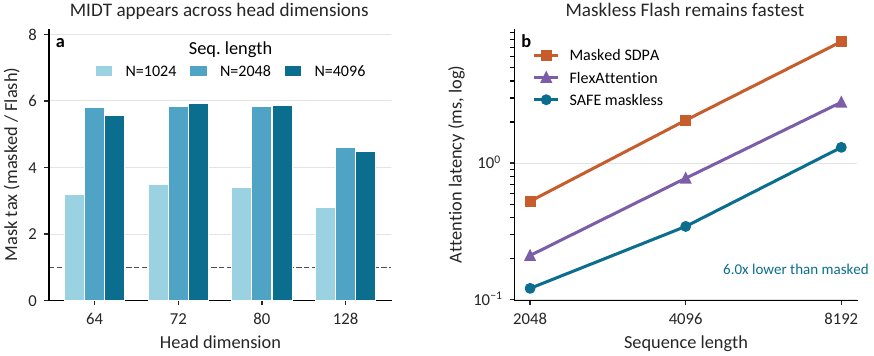}
  \caption{Mask-Induced Dispatch Tax. A redundant mask makes attention several
  times slower than the mask-free Flash path; the masked/Flash ratio grows with
  sequence length and persists across head dimensions.}
  \label{fig:supp_midt_tax}
\end{figure}

\paragraph{Kernel Dispatch.}
Table~\ref{tab:supp_dispatch} records the exact SDPA dispatch for a
head-dimension-72 call. Without a mask, SDPA dispatches to
\texttt{\_scaled\_dot\_product\_flash\_attention} and a
\texttt{pytorch\_flash::flash\_fwd\_kernel}. Passing an all-valid Boolean mask
switches dispatch to \texttt{\_scaled\_dot\_product\_efficient\_attention}
(a CUTLASS \texttt{fmha} kernel), making the call \(4.1\times\) to \(5.8\times\) slower
even though every key is visible. Compiled FlexAttention with an all-true block
mask narrows but does not close the gap (Fig.~\ref{fig:supp_midt_tax}).

\begin{table}[t]
\centering
\caption{Attention latency by mask and backend (head dim 72, bf16; ms,
median of 10 timed runs after 100 warm-ups).}
\label{tab:supp_dispatch}
\small
\setlength{\tabcolsep}{5pt}
\renewcommand{\arraystretch}{1.05}
\begin{tabular}{@{}l>{\columncolor{effbg}}c>{\columncolor{effbg}}c>{\columncolor{fidbg}}c@{}}
\toprule
\(N\) & no mask & all-valid mask & ratio \\
\midrule
2048 & 0.132 & 0.549 & \(4.1\times\) \\
4096 & 0.370 & 2.131 & \(5.8\times\) \\
\bottomrule
\end{tabular}
\end{table}

\paragraph{Third-Party Impact Surface.}
The tax is not specific to our pipeline. Table~\ref{tab:supp_midt_audit} audits
published implementations: the official \texttt{diffusers}
\texttt{LuminaAttnProcessor2\_0} passes an all-valid image self-attention mask
to SDPA unconditionally; the published RAS inherits the same masked path and
ships a manual \texttt{replace\_with\_flash\_attn} switch that bypasses SDPA via
an external \texttt{flash-attn} dependency, independently confirming both the
tax and its cost. Region/block-mask methods (SDiT-style) fall in the same
fallback tier, and guidance-map methods (S-CFG) read attention probabilities and
so cannot use FlashAttention at all. Measured against the no-mask Flash baseline
at \(N{=}4096\), an all-ones Boolean mask costs \(4.23\times\), a real text
padding mask \(4.28\times\), a block-diagonal region mask \(4.18\times\), and a
float additive bias \(3.45\times\).

\begin{table}[t]
\centering
\caption{MIDT impact surface across published implementations. ``Mask to
SDPA?'' is the default code path; ``Flash?'' is whether the FlashAttention fast
path is reached.}
\label{tab:supp_midt_audit}
\footnotesize
\setlength{\tabcolsep}{4pt}
\renewcommand{\arraystretch}{1.05}
\begin{tabular}{@{}l>{\columncolor{densebg}}l>{\columncolor{effbg}}l@{}}
\toprule
Implementation & Mask to SDPA? & Flash? \\
\midrule
diffusers Lumina-Next & yes (uncond.) & no (fallback) \\
RAS (default) & yes & no (fallback) \\
RAS (\texttt{flash} flag) & no (bypass) & yes (ext.\ dep.) \\
RAS on SD3 / PixArt & no (mask-free) & yes \\
SDiT-style block mask & yes (region) & no (fallback) \\
S-CFG (UNet) & N/A (no SDPA) & never \\
\rowcolor{safebg}
SAFE-DiT & elided (certified) & yes (automatic) \\
\bottomrule
\end{tabular}
\end{table}

\section{Mask-Exactness Matrix}
\label{sec:supp_exactness}

We characterize which attention masks the fast-path rewrite may remove. Using
an explicit FP32 reference \(\mathrm{softmax}(QK^\top/\sqrt{d}+m)V\), we measure,
for six mask families across \(N\in\{1024,2048,4096,8192\}\),
\(d_h\in\{64,72,128\}\), and three seeds: (i) the FP32 error of SAFE-DiT's
actual decision, (ii) the FP32 error of \emph{naively} eliding the mask
regardless of provenance, and (iii) the bf16 numerical floor of masked SDPA
against the FP32 reference. Table~\ref{tab:supp_exactness} reports the worst
case over all configurations.

\begin{table}[t]
\centering
\caption{Mask exactness (FP32 reference; worst-case max-abs error over all
\(N\), \(d_h\), seeds). Removable (row-constant) masks incur zero error under
SAFE-DiT; non-removable masks would incur large error if naively elided, so
SAFE-DiT preserves them. The bf16 column is the kernel-level numerical floor.}
\label{tab:supp_exactness}
\small
\setlength{\tabcolsep}{4pt}
\renewcommand{\arraystretch}{1.05}
\begin{tabular}{@{}ll>{\columncolor{effbg}}c>{\columncolor{rewbg}}c>{\columncolor{fidbg}}c@{}}
\toprule
Mask family & SAFE & SAFE & Naive & bf16 vs.\ \\
 & decision & err. & elide err. & FP32 \\
\midrule
\rowcolor{safebg}
All-valid Boolean      & elide    & \(0\)      & \(0\)      & \(1.6\text{e}{-}3\) \\
\rowcolor{safebg}
All-zero additive      & elide    & \(0\)      & \(0\)      & \(1.6\text{e}{-}3\) \\
\rowcolor{safebg}
Row-constant additive  & elide    & \(9\text{e}{-}8\) & \(9\text{e}{-}8\) & \(1.6\text{e}{-}3\) \\
\midrule
Text padding           & preserve & \(0\) & \(2.4\text{e}{-}1\) & \(1.5\text{e}{-}3\) \\
Block-local            & preserve & \(0\) & \(1.35\)            & \(3.9\text{e}{-}3\) \\
Finite non-uniform bias& preserve & \(0\) & \(6.0\text{e}{-}2\) & \(1.2\text{e}{-}3\) \\
\bottomrule
\end{tabular}
\end{table}

All-valid, all-zero, and row-constant masks are removable to machine precision
(\(\le 9\text{e}{-}8\) in FP32), whereas text-padding, block-local, and finite
non-uniform masks change the softmax and would introduce errors of
\(6\text{e}{-}2\) to \(1.35\) if elided. SAFE-DiT removes only the former and
preserves the latter, so correctness never depends on the elision firing.

\section{Full Statistics And FastCache Sweep}
\label{sec:supp_stats}

Table~\ref{tab:supp_ci} reports, for every method in the main comparison, the
ImageReward and its change versus Dense with a prompt-clustered paired
bootstrap (10{,}000 resamples; the three seeds of each prompt form one
cluster). AccelAes is included because it is the closest high-reward baseline:
both AccelAes and SAFE-DiT show significant positive reward gains over Dense,
so the main claim is matched quality at lower latency rather than uniqueness of
reward significance.

\begin{table}[t]
\centering
\caption{ImageReward and \(\Delta\)IR vs.\ Dense with 95\% paired-bootstrap
confidence intervals (DrawBench, 200 prompts \(\times\) 3 seeds). ``Sig.''
marks intervals excluding zero.}
\label{tab:supp_ci}
\small
\setlength{\tabcolsep}{4pt}
\renewcommand{\arraystretch}{1.02}
\begin{tabular}{@{}l>{\columncolor{rewbg}}c>{\columncolor{rewbg}}c>{\columncolor{fidbg}}cc@{}}
\toprule
Method & IR & \(\Delta\)IR & 95\% CI & Sig. \\
\midrule
AccelAes & 0.521 & \(+0.044\) & \([+0.026,+0.062]\) & \textbf{Y} \\
\rowcolor{safebg}
SAFE-DiT & 0.520 & \(+0.043\) & \([+0.022,+0.065]\) & \textbf{Y} \\
DPCache & 0.494 & \(+0.017\) & \([-0.001,+0.035]\) & N/A \\
RAS & 0.475 & \(-0.001\) & \([-0.008,+0.005]\) & N/A \\
TeaCache & 0.469 & \(-0.008\) & \([-0.014,-0.001]\) & \textbf{Y} \\
S-CFG & 0.443 & \(-0.034\) & \([-0.064,-0.005]\) & \textbf{Y} \\
TaylorSeer & 0.419 & \(-0.058\) & \([-0.085,-0.033]\) & \textbf{Y} \\
ClusCa & 0.384 & \(-0.093\) & \([-0.124,-0.064]\) & \textbf{Y} \\
FORA & 0.353 & \(-0.123\) & \([-0.161,-0.086]\) & \textbf{Y} \\
\(\Delta\)-DiT & 0.274 & \(-0.202\) & \([-0.244,-0.161]\) & \textbf{Y} \\
SDiT & 0.243 & \(-0.234\) & \([-0.275,-0.195]\) & \textbf{Y} \\
FastCache & \(-0.573\) & \(-1.050\) & \([-1.148,-0.955]\) & \textbf{Y} \\
\bottomrule
\end{tabular}
\end{table}

A direct paired SAFE-DiT versus AccelAes check gives a nonsignificant ImageReward
difference (SAFE-DiT \(-\) AccelAes \(=-0.001\), 95\% CI
\([-0.016,+0.014]\)). On the same 600 aligned prompt/seed timing samples,
SAFE-DiT is \(0.792\)\,s faster (95\% CI \([0.787,0.797]\)\,s), corresponding
to an AccelAes/SAFE-DiT speed ratio of \(1.225\times\) (95\% CI
\([1.223,1.226]\)).

\begin{figure}[t]
  \centering
  \includegraphics[width=\linewidth]{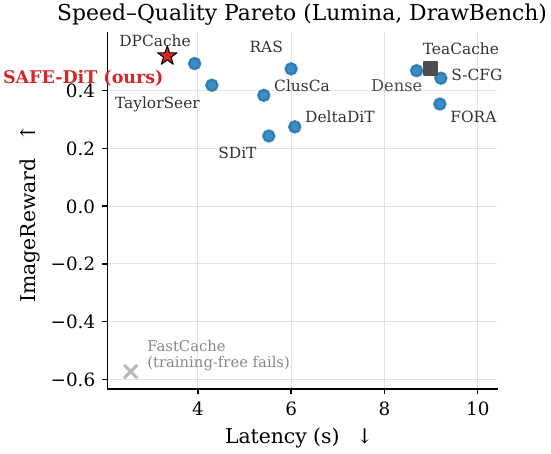}
  \caption{Speed and quality Pareto on Lumina-Next (DrawBench). SAFE-DiT occupies
  the top-left frontier (fast and high reward); FastCache is faster but low
  reward under the tested setting.}
  \label{fig:supp_pareto}
\end{figure}

\paragraph{FastCache Has No Fast High-Reward Operating Point.}
Under our Lumina-Next adaptation, we sweep FastCache's cache threshold on
24 prompts (Table~\ref{tab:supp_fastcache}). A threshold of 0 disables caching
and recovers Dense (no speedup); within the tested threshold range, every
accelerating setting drives IR negative. Thus, under this adaptation and sweep,
FastCache has no fast high-reward operating point, so we do not use it as the
primary high-reward comparator.

\begin{table}[t]
\centering
\caption{FastCache threshold sweep on Lumina-Next (24 prompts). Acceleration
and high reward are not obtained simultaneously in the tested range.}
\label{tab:supp_fastcache}
\small
\setlength{\tabcolsep}{6pt}
\renewcommand{\arraystretch}{1.02}
\begin{tabular}{@{}l>{\columncolor{rewbg}}c>{\columncolor{rewbg}}c@{}}
\toprule
Threshold & IR & \(\Delta\)IR vs.\ Dense \\
\midrule
0.00 (no cache) & 0.318 & \(+0.000\) \\
0.01 & 0.246 & \(-0.072\) \\
0.02 & \(-0.008\) & \(-0.326\) \\
0.05 (main table) & \(-0.639\) & \(-0.957\) \\
0.10 & \(-1.525\) & \(-1.843\) \\
0.20 & \(-2.016\) & \(-2.333\) \\
\bottomrule
\end{tabular}
\end{table}

\section{Fair Fast-Path Decomposition}
\label{sec:supp_decomp}

The main paper isolates the exact rewrite from the approximate scheduling on the
high-resolution stress suite. At \(2560^2\), keeping Lumina-Next's all-valid
mask (Dense-original) costs 209.3\,s and 94.1\,GB; applying only the exact
rewrite (Dense+FP) drops this to 62.6\,s and 24.7\,GB; adding query-only
scheduling is approximately latency-neutral (63.5\,s) after PCSP overhead; FFN
and cache reuse then bring SAFE-Core to 41.6\,s and 27.9\,GB, and SAFE-DiT+SW gives
41.1\,s and 27.9\,GB. At \(3072^2\), Dense-original and scheduling-without-elision
both run out of memory, Dense+FP completes at 126.7\,s / 26.9\,GB, and full
SAFE-DiT completes at 87.6\,s / 36.2\,GB.

An instrumented \(1024^2\) breakdown confirms the speedup lives in the denoiser
rather than in skipped peripheral stages: end-to-end / denoiser latency falls
from 9.32 / 8.89\,s (Dense) to 5.43 / 5.01\,s (mask elision) to 3.52 / 3.19\,s
(SAFE-DiT+SW), while transformer calls drop from 30 to 19. Text encoding,
VAE decoding, and post-processing stay small and stable (\(\sim\)0.03 / 0.13 /
0.02\,s), and the scheduler/PCSP overhead is 0.16 to 0.26\,s.

\section{High-Resolution Frontier}
\label{sec:supp_frontier}

Table~\ref{tab:supp_frontier} gives the full numeric frontier on Lumina-Next,
and Table~\ref{tab:supp_frontier_base} adds the strongest acceleration
baselines. Temporal caches skip steps but still materialize full attention per
computed step, so their peak memory stays near Dense; consequently at
\(2560^2\) Dense, DPCache, and TeaCache all run out of memory while only
SAFE-DiT, which reduces per-step token computation, completes.

\begin{table}[t]
\centering
\caption{Lumina-Next speed and memory frontier (SAFE-DiT vs.\ Dense). At
\(3072^2\) Dense is OOM.}
\label{tab:supp_frontier}
\small
\setlength{\tabcolsep}{4pt}
\renewcommand{\arraystretch}{1.02}
\begin{tabular}{@{}l>{\columncolor{effbg}}c>{\columncolor{effbg}}c>{\columncolor{effbg}}c@{}}
\toprule
Res. & Speedup & Dense mem.\ (GB) & SAFE mem.\ (GB) \\
\midrule
\(1024^2\) & \(2.69\times\) & 11.3 & 11.7 \\
\(1536^2\) & \(3.50\times\) & 20.2 & 15.0 \\
\(2048^2\) & \(4.41\times\) & 44.1 & 19.7 \\
\rowcolor{safebg}
\(2560^2\) & \(5.09\times\) & 94.1 & 27.9 \\
\(3072^2\) & OOM\(\to\)runs & N/A & 36.2 \\
\bottomrule
\end{tabular}
\end{table}

\begin{table}[t]
\centering
\caption{High-resolution comparison with acceleration baselines (latency\,s /
peak GB). Temporal caches do not reduce memory and OOM at \(2560^2\).}
\label{tab:supp_frontier_base}
\small
\setlength{\tabcolsep}{4pt}
\renewcommand{\arraystretch}{1.02}
\begin{tabular}{@{}l>{\columncolor{effbg}}c>{\columncolor{effbg}}c@{}}
\toprule
Method & \(2048^2\) & \(2560^2\) \\
\midrule
Dense & 92.9 / 44.1 & OOM \\
TeaCache & 86.9 / 44.4 & OOM \\
DPCache & 31.1 / 44.7 & OOM \\
\rowcolor{safebg}
SAFE-DiT & \textbf{21.3 / 21.3} & \textbf{41.1 / 27.9} \\
\bottomrule
\end{tabular}
\end{table}

\section{A800 Cross-Architecture Evaluation}
\label{sec:supp_a800}

We additionally evaluate SAFE-DiT on an NVIDIA A800-SXM4-80GB GPU
(Ampere; PyTorch 2.8.0+cu128; CUDA runtime 12.8) to test whether MIDT and the
resulting acceleration persist outside the RTX PRO 6000 Blackwell stack. This
experiment is not intended to match Blackwell speedups; it checks whether
redundant masks still change SDPA dispatch, and whether Dense+FP and SAFE-Core
still provide stable gains.

Table~\ref{tab:supp_a800_kernel} shows that the A800 stack dispatches no-mask
SDPA to FlashAttention, but an all-valid mask to the memory-efficient backend.
FlexAttention uses Triton kernels and is faster than the all-valid mask path but
still slower than the no-mask Flash path at all tested sequence lengths.

\begin{table}[t]
\centering
\caption{A800 kernel microbenchmark. Entries report median latency in ms and
the observed dispatch.}
\label{tab:supp_a800_kernel}
\footnotesize
\setlength{\tabcolsep}{4pt}
\renewcommand{\arraystretch}{1.03}
\begin{tabular}{@{}r>{\columncolor{effbg}}c>{\columncolor{fidbg}}c>{\columncolor{rewbg}}c@{}}
\toprule
\(N\) & No mask & All-valid mask & FlexAttention \\
\midrule
2048 & 0.268 / Flash & 0.919 / mem-eff. & 0.863 / Triton \\
4096 & 0.805 / Flash & 3.425 / mem-eff. & 2.739 / Triton \\
8192 & 3.004 / Flash & 10.830 / mem-eff. & 8.344 / Triton \\
\bottomrule
\end{tabular}
\end{table}

Table~\ref{tab:supp_a800_matched} reports the matched end-to-end comparison on
20 DrawBench prompts with one fixed seed. Dense+FP removes the mask-dispatch
tax, and SAFE-Core adds a further scheduling gain of about \(1.55\times\) to \(1.57\times\)
over Dense+FP from \(1024^2\) to \(2048^2\). SAFE-DiT+SW has nearly identical
latency and memory to SAFE-Core, confirming that SW-CFG is not a compute
bottleneck on this stack.

\begin{table*}[t]
\centering
\caption{A800 matched comparison on 20 prompts \(\times\) 1 seed. Latency is
the median wall-clock time in seconds; memory is peak allocated GB.}
\label{tab:supp_a800_matched}
\scriptsize
\setlength{\tabcolsep}{4pt}
\renewcommand{\arraystretch}{1.02}
\begin{tabular}{@{}rl>{\columncolor{densebg}}c>{\columncolor{effbg}}c>{\columncolor{effbg}}c>{\columncolor{rewbg}}c>{\columncolor{rewbg}}c@{}}
\toprule
Res. & Config & OK/OOM/Error & Latency & Memory & Speedup vs Dense & Speedup vs Dense+FP \\
\midrule
\(1024^2\) & Dense & 20/0/0 & 14.826 & 11.342 & \(1.000\times\) & \(0.594\times\) \\
\(1024^2\) & Dense+FP & 20/0/0 & 8.804 & 11.342 & \(1.684\times\) & \(1.000\times\) \\
\(1024^2\) & SAFE-Core & 20/0/0 & 5.694 & 11.529 & \(2.604\times\) & \(1.546\times\) \\
\rowcolor{safebg}
\(1024^2\) & SAFE-DiT+SW & 20/0/0 & 5.731 & 11.529 & \(2.587\times\) & \(1.536\times\) \\
\midrule
\(1536^2\) & Dense & 20/0/0 & 50.873 & 20.179 & \(1.000\times\) & \(0.454\times\) \\
\(1536^2\) & Dense+FP & 20/0/0 & 23.078 & 14.534 & \(2.204\times\) & \(1.000\times\) \\
\(1536^2\) & SAFE-Core & 20/0/0 & 14.691 & 14.955 & \(3.463\times\) & \(1.571\times\) \\
\rowcolor{safebg}
\(1536^2\) & SAFE-DiT+SW & 20/0/0 & 14.697 & 14.955 & \(3.461\times\) & \(1.570\times\) \\
\midrule
\(2048^2\) & Dense & 20/0/0 & 140.261 & 44.068 & \(1.000\times\) & \(0.371\times\) \\
\(2048^2\) & Dense+FP & 20/0/0 & 52.007 & 19.000 & \(2.697\times\) & \(1.000\times\) \\
\(2048^2\) & SAFE-Core & 20/0/0 & 33.186 & 19.747 & \(4.227\times\) & \(1.567\times\) \\
\rowcolor{safebg}
\(2048^2\) & SAFE-DiT+SW & 20/0/0 & 33.422 & 19.747 & \(4.197\times\) & \(1.556\times\) \\
\bottomrule
\end{tabular}
\end{table*}

Table~\ref{tab:supp_a800_stress} gives the A800 stress frontier. Dense runs out
of memory at both \(2560^2\) and \(3072^2\). Dense+FP completes both settings,
and SAFE-Core / SAFE-DiT+SW further reduce latency by \(1.56\times\) at
\(2560^2\) and \(1.43\times\) at \(3072^2\) relative to Dense+FP.

\begin{table}[t]
\centering
\caption{A800 stress frontier on 20 prompts \(\times\) 1 seed. Latency and
memory are reported only for completed runs.}
\label{tab:supp_a800_stress}
\footnotesize
\setlength{\tabcolsep}{4pt}
\renewcommand{\arraystretch}{1.02}
\resizebox{\columnwidth}{!}{%
\begin{tabular}{@{}rl>{\columncolor{densebg}}c>{\columncolor{effbg}}c>{\columncolor{effbg}}c>{\columncolor{rewbg}}c@{}}
\toprule
Res. & Config & OK/OOM & Latency & Memory & Speedup vs FP \\
\midrule
\(2560^2\) & Dense & 0/20 & N/A & N/A & N/A \\
\(2560^2\) & Dense+FP & 20/0 & 104.869 & 24.743 & \(1.000\times\) \\
\(2560^2\) & SAFE-Core & 20/0 & 67.222 & 25.915 & \(1.560\times\) \\
\rowcolor{safebg}
\(2560^2\) & SAFE-DiT+SW & 20/0 & 67.402 & 25.915 & \(1.556\times\) \\
\midrule
\(3072^2\) & Dense & 0/20 & N/A & N/A & N/A \\
\(3072^2\) & Dense+FP & 20/0 & 228.000 & 26.931 & \(1.000\times\) \\
\(3072^2\) & SAFE-Core & 20/0 & 159.126 & 33.451 & \(1.433\times\) \\
\rowcolor{safebg}
\(3072^2\) & SAFE-DiT+SW & 20/0 & 159.074 & 33.451 & \(1.433\times\) \\
\bottomrule
\end{tabular}
}
\end{table}

\section{Cross-Backbone Generalization}
\label{sec:supp_xbackbone}

Table~\ref{tab:supp_xbackbone} expands the main-paper cross-backbone summary.
Speedup tracks the self-attention mask: the masked backbone (Lumina-Next) gains
the most and scales with resolution because the rewrite and scheduling compound,
while mask-free backbones obtain scheduling-only gains that depend on the model
(large on the step-heavy FLUX, smaller on PixArt-\(\Sigma\)). Quality is stable
for Lumina-Next, SD3-Medium, FLUX.1-dev, and the selected PixArt-\(\Sigma\)
\(2048^2\) operating point. PixArt-\(\Sigma\) shows larger reward drops at
\(1024^2\) and \(3072^2\), which we attribute to relying on approximate
scheduling alone without an exact mask-elision anchor under the tested
configuration.

\begin{table}[t]
\centering
\caption{Cross-backbone results (paired prompts/seeds). Self-attn indicates
whether the backbone triggers MIDT.}
\label{tab:supp_xbackbone}
\footnotesize
\setlength{\tabcolsep}{3.5pt}
\renewcommand{\arraystretch}{1.04}
\begin{tabular}{@{}lll>{\columncolor{effbg}}c>{\columncolor{fidbg}}c>{\columncolor{fidbg}}c@{}}
\toprule
Backbone & Res. & Self-attn & Speed. & \(\Delta\)IR & LPIPS \\
\midrule
\rowcolor{safebg}
Lumina-Next & \(1024^2\) & masked & \(2.69\times\) & \(+0.043\) & 0.119 \\
\rowcolor{safebg}
Lumina-Next & \(2560^2\) & masked & \(5.09\times\) & N/A & N/A \\
SD3-Medium & \(1024^2\) & mask-free & \(1.30\times\) & \(+0.004\) & 0.026 \\
FLUX.1-dev & \(1024^2\) & mask-free & \(5.14\times\) & \(-0.009\) & 0.154 \\
PixArt-\(\Sigma\) & \(1024^2\) & mask-free & \(1.76\times\) & \(-0.188\) & N/A \\
PixArt-\(\Sigma\) & \(2048^2\) & mask-free & \(1.87\times\) & \(+0.028\) & 0.307 \\
PixArt-\(\Sigma\) & \(3072^2\) & mask-free & \(1.55\times\) & \(-0.116\) & 0.360 \\
\bottomrule
\end{tabular}
\end{table}

\begin{figure}[t]
  \centering
  \includegraphics[width=\linewidth]{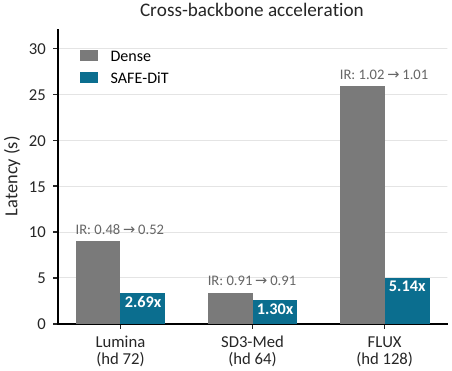}
  \caption{Cross-backbone speedup. The masked backbone (Lumina-Next) gains most
  and scales with resolution; mask-free backbones obtain model-dependent
  scheduling-only gains.}
  \label{fig:supp_xbackbone}
\end{figure}

\section{Ablations And Sensitivity}
\label{sec:supp_ablation}

\begin{figure}[t]
  \centering
  \includegraphics[width=\linewidth]{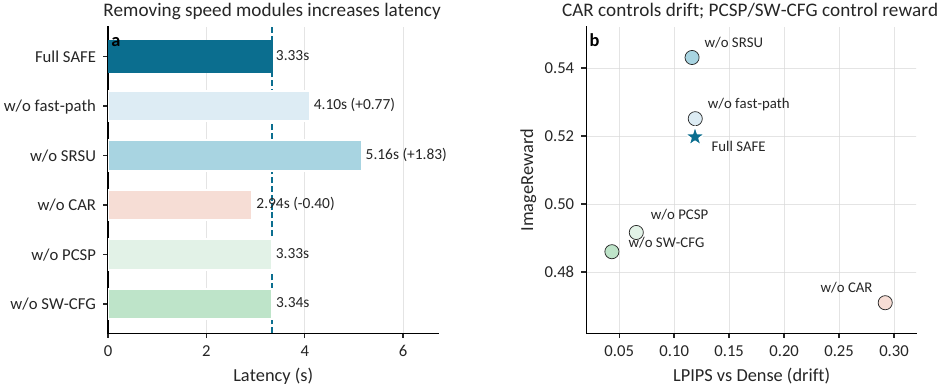}
  \caption{Module roles. Left: removing a speed module (fast-path, SRSU)
  inflates latency, placing those variants at slower conservative operating
  points. Right: in the reward and drift plane, removing CAR drives drift (LPIPS)
  while removing PCSP/SW-CFG lowers reward.}
  \label{fig:supp_ablation}
\end{figure}

\paragraph{Hyperparameter Sweeps.}
Table~\ref{tab:supp_sweeps} sweeps the sensitive-token ratio
(\(\rho\)\,/\,skip ratio), the dense warm-up length, and the anchor interval on
a development set with 50 to 60 prompts (IR is on this set's scale). Each knob trades
latency for fidelity, and the released default sits at the speed and fidelity knee
rather than at an extreme.

\begin{table}[t]
\centering
\caption{Hyperparameter sweeps (development set). Higher skip ratio and longer
anchor interval are faster but drift; longer warm-up improves fidelity at a
latency cost.}
\label{tab:supp_sweeps}
\small
\setlength{\tabcolsep}{5pt}
\renewcommand{\arraystretch}{1.0}
\begin{tabular}{@{}l>{\columncolor{rewbg}}c>{\columncolor{fidbg}}c@{}}
\toprule
Setting & IR & LPIPS \\
\midrule
skip ratio 0.3 & 0.765 & 0.105 \\
skip ratio 0.5 (default) & 0.747 & 0.103 \\
skip ratio 0.7 & 0.725 & 0.109 \\
\midrule
warm-up 3 & 0.777 & 0.145 \\
warm-up 7 & 0.733 & 0.078 \\
warm-up 9 & 0.743 & 0.061 \\
\midrule
anchor interval 1 & 0.747 & 0.103 \\
anchor interval 4 & 0.759 & 0.113 \\
anchor interval 8 & 0.753 & 0.173 \\
\bottomrule
\end{tabular}
\end{table}

\paragraph{CAR Drift.}
Table~\ref{tab:supp_car} lengthens the anchor-refresh interval. Without context
anchor refresh, cached context states drift and LPIPS rises sharply
(\(0.10\to0.28\)); short refresh intervals keep drift bounded. This isolates
CAR as a drift-control rather than a speed component.

\begin{table}[t]
\centering
\caption{CAR drift vs.\ refresh interval (development set). Weakening CAR is
faster but drifts.}
\label{tab:supp_car}
\small
\setlength{\tabcolsep}{6pt}
\renewcommand{\arraystretch}{1.0}
\begin{tabular}{@{}l>{\columncolor{rewbg}}c>{\columncolor{fidbg}}c@{}}
\toprule
Refresh interval & IR & LPIPS \\
\midrule
1 & 0.817 & 0.102 \\
4 & 0.823 & 0.111 \\
8 & 0.813 & 0.169 \\
16 & 0.789 & 0.235 \\
none & 0.779 & 0.283 \\
\bottomrule
\end{tabular}
\end{table}

\begin{figure}[t]
  \centering
  \includegraphics[width=\linewidth]{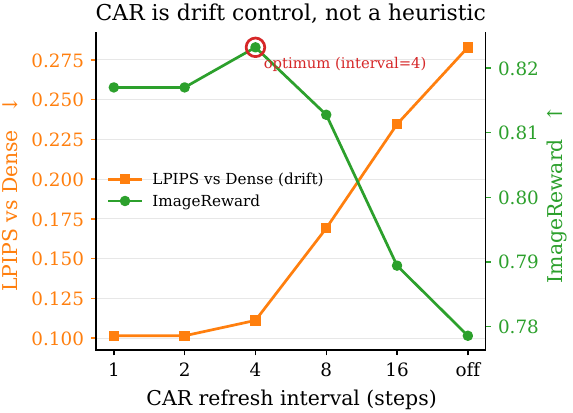}
  \caption{Context-state drift over denoising steps. Without CAR, cached context
  states drift monotonically; anchor refreshes reset the error, bounding drift.}
  \label{fig:supp_car}
\end{figure}

\section{PCSP Is Content-Driven}
\label{sec:supp_pcsp}

\begin{figure}[t]
  \centering
  \includegraphics[width=\linewidth]{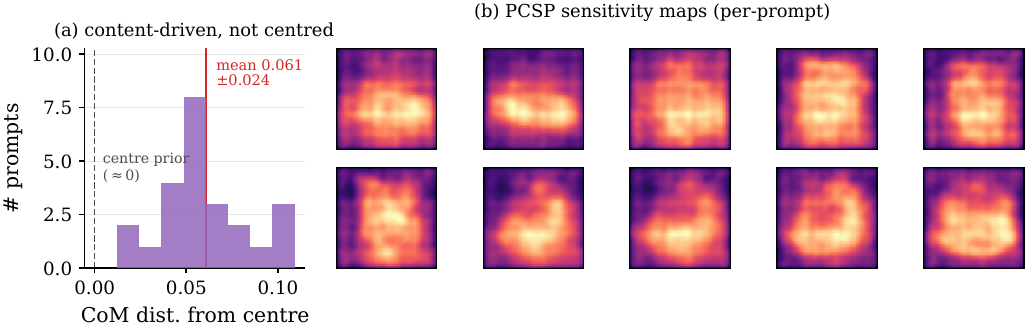}
  \caption{PCSP sensitivity. (a) The partition's center of mass is displaced from
  the image center, not centered. (b) Per-prompt sensitivity maps localize on
  prompt-relevant regions.}
  \label{fig:supp_pcsp}
\end{figure}

PCSP partitions tokens from early image-to-text attention rather than from a
fixed spatial prior. Across captured prompts (Fig.~\ref{fig:supp_pcsp}), the
sensitivity-map center of mass is displaced from the image center by
\(0.061\pm0.024\) and varies with the prompt, confirming a content-driven
partition rather than a center bias.

\section{Perceptual Evaluation}
\label{sec:supp_perceptual}

\paragraph{Blind VLM Judgments.}
On a 100-prompt paired DrawBench subset with randomized left/right order,
GPT-5 scored SAFE-DiT versus Dense at 37 wins, 32 ties, and 31 losses
(tie-excluded win rate 54.4\%, binomial \(p{=}0.54\)); GPT-4o scored 14 / 82 / 4,
with most pairs tied. Both support quality \emph{parity} rather than
improvement, so we phrase the perceptual claim as preservation.

\begin{figure}[t]
  \centering
  \includegraphics[width=\linewidth]{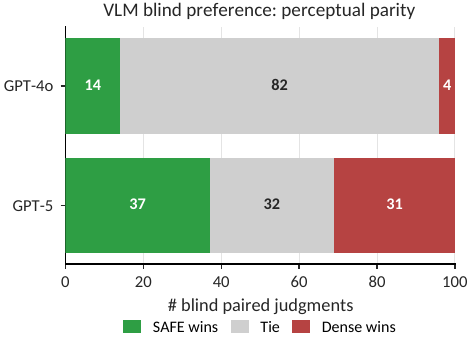}
  \caption{Blind VLM win/tie/loss for SAFE-DiT vs.\ Dense. Both judges place most
  pairs at parity, supporting preservation rather than improvement.}
  \label{fig:supp_vlm}
\end{figure}

\paragraph{T2I-CompBench.}
On 300 validation prompts per category, SAFE-Core matches Dense across
categories (attribute-binding average 0.406 vs.\ 0.404), while SW-CFG raises the
attribute-binding average to 0.419 (color 0.481\,$\to$\,0.509, texture
0.421\,$\to$\,0.432) without changing non-spatial relations
(Table~\ref{tab:supp_compbench}). This shows that the SW-CFG module mainly
acts on prompt alignment, while SAFE-Core isolates acceleration fidelity.

\begin{table}[t]
\centering
\caption{T2I-CompBench alignment on Lumina-Next.}
\label{tab:supp_compbench}
\small
\setlength{\tabcolsep}{5pt}
\renewcommand{\arraystretch}{1.02}
\begin{tabular}{@{}l>{\columncolor{densebg}}c>{\columncolor{effbg}}c>{\columncolor{safebg}}c@{}}
\toprule
Category & Dense & SAFE-Core & SAFE-DiT+SW \\
\midrule
Color & 0.481 & 0.481 & \textbf{0.509} \\
Shape & 0.310 & 0.312 & \textbf{0.317} \\
Texture & 0.421 & 0.425 & \textbf{0.432} \\
Non-spatial & 0.303 & 0.303 & 0.303 \\
\midrule
Attr.-binding avg. & 0.404 & 0.406 & \textbf{0.419} \\
\bottomrule
\end{tabular}
\end{table}

\paragraph{GenEval.}
We further evaluate object-centric compositional alignment with the official
GenEval detector-based protocol: 553 prompts, four images per prompt, and the
standard task-average score over single object, two object, counting, colors,
position, and color attribution. SAFE-Core remains matched to Dense in the
overall score (0.449 vs.\ 0.449), while SAFE-DiT+SW gives a small alignment gain
to 0.458, mainly from counting, position, and color attribution
(Table~\ref{tab:supp_geneval}). The low absolute position scores are shared by
all three configurations, so we treat spatial-relation failures as a backbone
limitation rather than an acceleration artifact. The table is therefore used as
a compact sanity check: acceleration leaves the detector-based failure modes
nearly unchanged, while SW-CFG gives only a modest alignment shift.

\begin{table}[H]
\centering
\caption{GenEval object-centric alignment on Lumina-Next.}
\label{tab:supp_geneval}
\scriptsize
\setlength{\tabcolsep}{2pt}
\renewcommand{\arraystretch}{1.02}
\resizebox{\columnwidth}{!}{%
\begin{tabular}{@{}l>{\columncolor{rewbg}}c>{\columncolor{effbg}}c>{\columncolor{effbg}}c>{\columncolor{fidbg}}c>{\columncolor{fidbg}}c>{\columncolor{fidbg}}c>{\columncolor{fidbg}}c>{\columncolor{fidbg}}c>{\columncolor{fidbg}}c@{}}
\toprule
Method & Overall & Img. ok & Prompt ok & Single & Two &
Count & Color & Pos. & Attr. \\
\midrule
Dense & 0.449 & 43.17 & 58.95 & \textbf{90.94} & 43.43 &
38.44 & \textbf{72.61} & 8.00 & 16.00 \\
SAFE-Core & 0.449 & 43.17 & 60.04 & \textbf{90.94} & 44.19 &
37.81 & 71.81 & 8.00 & 16.50 \\
\rowcolor{safebg}
SAFE-DiT+SW & \textbf{0.458} & \textbf{44.03} & \textbf{60.58} & 90.62 &
\textbf{44.70} & \textbf{42.19} & 71.54 & \textbf{8.50} & \textbf{17.25} \\
\bottomrule
\end{tabular}
}
\end{table}

\section{Qualitative Comparisons}
\label{sec:supp_qual}

\begin{figure*}[t]
  \centering
  \includegraphics[width=0.875\textwidth]{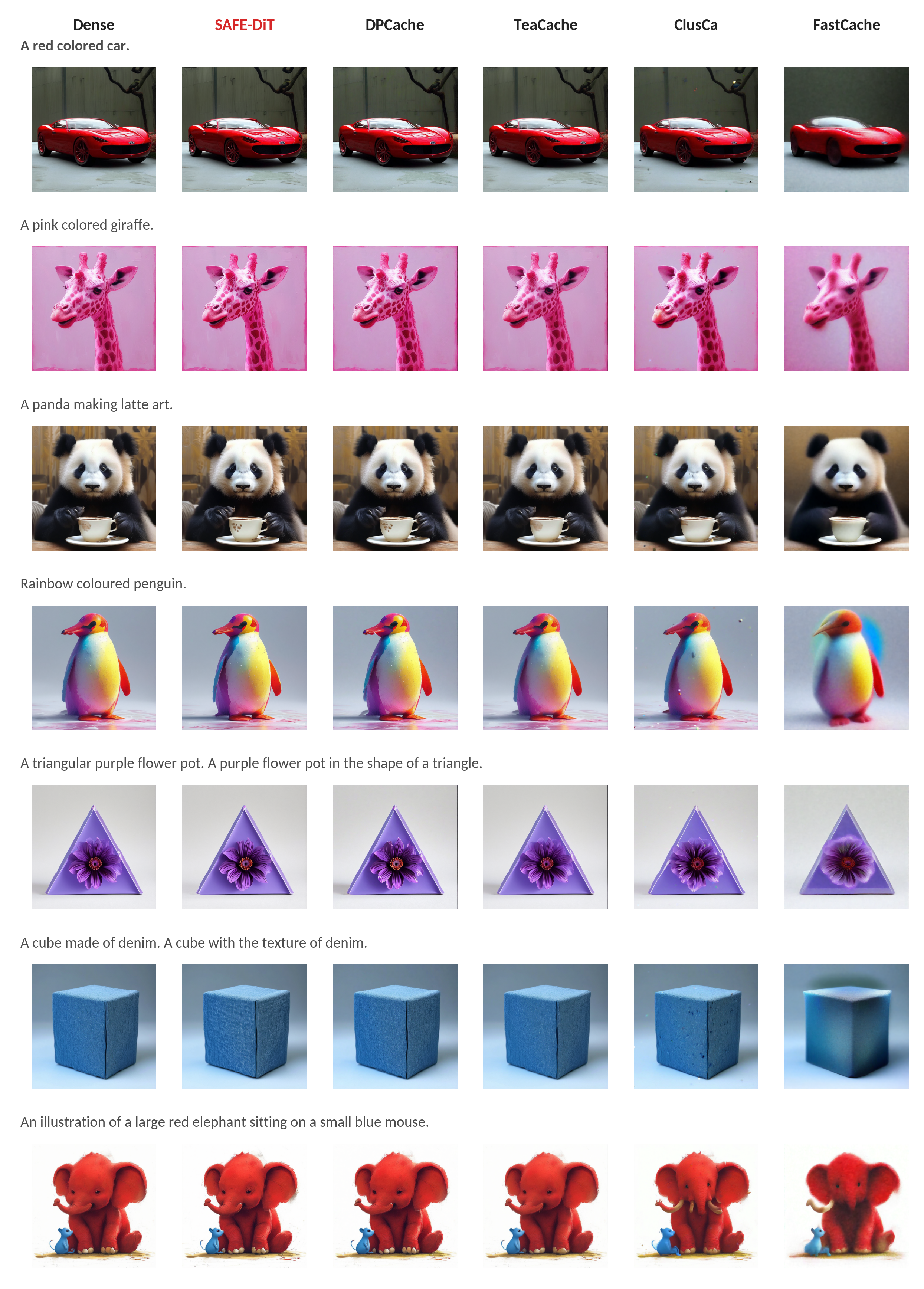}
  \caption{Paired qualitative comparison (matched prompt/seed). Columns: Dense,
  SAFE-DiT, DPCache, TeaCache, ClusCa, FastCache. SAFE-DiT preserves the Dense
  composition and detail; FastCache (rightmost) visibly degrades.}
  \label{fig:supp_grid}
\end{figure*}

Figure~\ref{fig:supp_grid} compares SAFE-DiT with Dense and four acceleration
baselines on Lumina-Next under matched prompt and seed. SAFE-DiT tracks the
Dense reference, while the strongest-quality baselines remain close and FastCache
degrades visibly.

\FloatBarrier

\begin{table*}[!t]
\centering
\caption{
Blinded human-study results after quality control.
``Target'' and ``reference'' denote the first and second methods,
respectively, in the comparison column.
Preference rates and bootstrap confidence intervals exclude ties.
The lower confidence bound is the one-sided 95\% Wilson bound
computed over decisive prompt-level outcomes.
}
\label{tab:supp_human}
\small
\setlength{\tabcolsep}{4pt}
\renewcommand{\arraystretch}{1.05}
\resizebox{\textwidth}{!}{%
\begin{tabular}{@{}ll
>{\columncolor{effbg}}c
>{\columncolor{densebg}}c
>{\columncolor{effbg}}c
>{\columncolor{rewbg}}c
>{\columncolor{fidbg}}c
>{\columncolor{fidbg}}l@{}}
\toprule
Comparison & Criterion &
Target wins & Ties & Ref. wins &
Tie-excl. pref. & Bootstrap 95\% CI &
Wilson LCB / conclusion \\
\midrule
SAFE-Core vs.\ Dense+FP
& Visual quality
& 4 & 196 & 0
& 1.000 & [\(1.000, 1.000\)]
& 0.597; non-inferior \\

SAFE-Core vs.\ Dense+FP
& Prompt alignment
& 1 & 199 & 0
& 1.000 & [\(1.000, 1.000\)]
& 0.270; descriptive only \\

SAFE-DiT+SW vs.\ SAFE-Core
& Visual quality
& 4 & 41 & 5
& 0.444 & [\(0.111, 0.800\)]
& 0.218; no preference claim \\

SAFE-DiT+SW vs.\ SAFE-Core
& Prompt alignment
& 1 & 48 & 1
& 0.500 & [\(0.000, 1.000\)]
& 0.121; no preference claim \\
\bottomrule
\end{tabular}
}
\end{table*}

\section{Human Study Protocol and Non-Inferiority Analysis}
\label{sec:supp_human}

To complement the automatic metrics and VLM-based evaluation, we conducted
a blinded paired human study that separates acceleration fidelity from the
effect of the SW-CFG guidance module. The primary comparison evaluates
Dense+FP against SAFE-Core and tests whether the exact fast-path rewrite and
approximate spatial scheduling preserve perceived visual quality under matched
prompts, seeds, samplers, numbers of function evaluations, and \(1024^2\)
resolution. The secondary comparison evaluates SAFE-Core against
SAFE-DiT+SW and examines whether SW-CFG changes prompt alignment or perceived
visual quality. We analyze these comparisons separately rather than combining
acceleration fidelity and guidance effects into a single preference endpoint.

The study was implemented using a public Hugging Face Space with a private
result store. Each trial displayed the full text prompt and two anonymized
images labelled only as Image A and Image B. The left--right order was
randomized independently for each rater assignment. Method names, filenames,
runtime measurements, and automatic metric values were hidden from the
raters. For each pair, raters first selected which image better matched the
prompt and then selected which image had better visual quality. A tie option
was available for both questions. We recruited adult raters who were not
involved in the project and stored only anonymous participant identifiers.

Five raters each completed 160 trials, yielding 800 responses in total. These
comprised 200 primary pairs evaluated by three raters
(\(600\) judgments), 50 stratified secondary pairs evaluated by three raters
(\(150\) judgments), and 50 attention-check trials. Before data collection,
we fixed the prompt identifiers, random seeds, image files, rater-assignment
file, quality-control rules, and non-inferiority margin.

Quality control was applied at the rater level using two types of attention
checks. Identical-image checks required a tie response for both visual quality
and prompt alignment. For original-versus-blurred checks, the visual-quality
response was required to select the sharp original image; for prompt
alignment, either the original image or a tie was accepted because the blur
manipulation primarily tests visual quality. One rater was excluded after
failing three identical-image checks. After exclusion, the valid dataset
contained 600 formal judgments covering all 250 prespecified formal pairs.
Of these pairs, 100 retained three valid ratings and 150 retained two valid
ratings.

We first aggregate valid ratings independently for each prompt-level image
pair. The majority outcome is categorized as a target-method win, a tie, or a
reference-method win. A split vote with no strict majority between the two
methods is counted as a tie. All subsequent statistical analyses operate on
these prompt-level outcomes; individual rater judgments are not treated as
independent observations.

For a comparison with \(N_{\mathrm{target}}\) target-method wins and
\(N_{\mathrm{reference}}\) reference-method wins, we define the tie-excluded
preference rate as
\[
p_{\mathrm{target}}
=
\frac{N_{\mathrm{target}}}
     {N_{\mathrm{target}} + N_{\mathrm{reference}}}.
\]
This quantity is conditional on a prompt-level comparison being decisive.
Because ties are excluded from its denominator, we report the numbers of
wins, ties, and losses alongside every preference rate.

We report two complementary uncertainty summaries. First, we compute a
two-sided 95\% nonparametric bootstrap confidence interval by resampling the
prompt-level majority outcomes with replacement and recomputing the
tie-excluded preference rate. Bootstrap replicates containing no decisive
outcome are omitted because the tie-excluded rate is undefined in those
replicates. Second, we compute a one-sided 95\% Wilson lower confidence bound
over the decisive prompt-level outcomes. The Wilson bound is used for the
prespecified non-inferiority decision, while the bootstrap interval is
reported as a descriptive uncertainty summary.

For the primary visual-quality endpoint, we use a 5\% non-inferiority margin
relative to equal preference and test
\[
H_0: p_{\mathrm{SAFE}} \leq 0.45
\qquad\text{against}\qquad
H_1: p_{\mathrm{SAFE}} > 0.45.
\]
SAFE-Core is declared non-inferior to Dense+FP in perceived visual quality
only when the one-sided 95\% Wilson lower confidence bound exceeds \(0.45\).
Endpoints that do not satisfy this criterion are reported descriptively using
their prompt-level wins, ties, losses, preference rates, and confidence
intervals.

For the secondary comparison involving SW-CFG, we do not apply the primary
non-inferiority test. A human preference claim for prompt alignment would
require the two-sided prompt-level confidence interval to lie entirely above
\(0.5\). Otherwise, the result is interpreted descriptively and no preference
claim is made.

As shown in Table~\ref{tab:supp_human}, SAFE-Core received four prompt-level
visual-quality wins, 196 ties, and no losses against Dense+FP. Its one-sided
95\% Wilson lower confidence bound is \(0.597\), which exceeds the
prespecified non-inferiority threshold of \(0.45\). SAFE-Core therefore
satisfies the stated 5\% non-inferiority criterion for perceived visual
quality. This result does not establish visual superiority because the
comparison is overwhelmingly tie dominated.

The secondary comparison between SAFE-DiT+SW and SAFE-Core is also dominated
by ties. Its prompt-alignment confidence interval includes \(0.5\), and the
study therefore does not support a human preference claim for SW-CFG.
Overall, the human evaluation supports perceptual non-inferiority of
SAFE-Core to Dense+FP under the evaluated configuration, rather than evidence
that either SAFE-Core or SAFE-DiT+SW is perceptually superior.

\end{document}